\documentclass[11pt, a4paper]{article}

\usepackage[english]{babel} 
\usepackage{booktabs} 
\usepackage{comment} 
\usepackage[utf8]{inputenc} 
\usepackage[T1]{fontenc} 

\usepackage{amsmath,amsfonts,amsthm}
\usepackage{tikz} 
\usepackage{tikz-cd}
\usetikzlibrary{positioning,arrows} 
\usetikzlibrary{decorations.pathreplacing} 
\usepackage{tikzsymbols} 
\usepackage{blindtext} 
\usepackage[colorlinks=true, linkcolor=blue, citecolor=blue, urlcolor=blue]{hyperref}

\usepackage{geometry}
\usepackage{lscape}
\usepackage{pdflscape}  
\usepackage{graphicx}    
\usepackage{booktabs} 
\usepackage{multirow}
\usepackage{booktabs,tabularx,amssymb}
\newcolumntype{C}{>{\centering\arraybackslash}m{0.55cm}}
\usepackage{rotating}

\usepackage{subcaption}

\usepackage{caption}
\usepackage{algorithm}
\usepackage{algpseudocode}

\usepackage{geometry} 

\usepackage{setspace} 
\usepackage[T1]{fontenc} 
\usepackage[utf8]{inputenc} 

\usepackage{XCharter} 

\usepackage{indentfirst}

\usepackage{fancyhdr} 
\usepackage[natbibapa]{apacite}

\newcommand{\keywords}[1]{%
  \par\addvspace{\baselineskip}%
  \noindent\textbf{Keywords: }{\begingroup\def\and{, }#1\endgroup}%
}
\newcommand{\figref}[1]{\hyperref[#1]{Fig.~\ref*{#1}}}
\newcommand{\tabref}[1]{\hyperref[#1]{Table~\ref*{#1}}}

\title{Prior-informed optimization of treatment recommendation via bandit algorithms trained on large language model-processed historical records}

\author{
    Saman Nessari, 
    Ali Bozorgi-Amiri\thanks{Corresponding author.\\Email addresses: 
    \href{mailto:saman.nesari@ut.ac.ir}{\textit{saman.nesari@ut.ac.ir}} (S. Nessari), 
    \href{mailto:alibozorgi@ut.ac.ir}{\textit{alibozorgi@ut.ac.ir}} (A. Bozorgi-Amiri)}, 
    \\
    School of Industrial Engineering, College of Engineering, University of Tehran, Tehran, Iran
}

\date{}

\begin{document}
	\maketitle
\setcounter{page}{1} 

\begin{abstract}

    \noindent Current medical practice depends on standardized treatment frameworks and empirical methodologies that neglect individual patient variations, leading to suboptimal health outcomes. We develop a comprehensive system integrating Large Language Models (LLMs), Conditional Tabular Generative Adversarial Networks (CTGAN), T-learner counterfactual models, and contextual bandit approaches to provide customized, data-informed clinical recommendations. The approach utilizes LLMs to process unstructured medical narratives into structured datasets (93.2\% accuracy), uses CTGANs to produce realistic synthetic patient data (55\% accuracy via two-sample verification), deploys T-learners to forecast patient-specific treatment responses (84.3\% accuracy), and integrates prior-informed contextual bandits to enhance online therapeutic selection by effectively balancing exploration of new possibilities with exploitation of existing knowledge. Testing on stage III colon cancer datasets revealed that our KernelUCB approach obtained 0.60-0.61 average reward scores across 5,000 rounds, exceeding other reference methods. This comprehensive system overcomes cold-start limitations in online learning environments, improves computational effectiveness, and constitutes notable progress toward individualized medicine adapted to specific patient characteristics.


\end{abstract}


\section{Introduction}

Current healthcare delivery largely employs an experimental treatment model, applying generic clinical protocols that disregard patient-specific factors and individual differences. This standardized methodology requires patients to experience multiple treatment cycles that frequently lack efficacy, thereby prolonging patient discomfort, amplifying medical costs, and delaying positive health outcomes across diverse specialties ranging from mental health to cancer treatment. Although Electronic Health Records (EHRs) have become ubiquitous in healthcare settings, therapeutic decision-making continues to operate independently from previously recorded treatment outcomes and crucial clinical insights embedded within unstructured clinical documentation that conventional analytical methods cannot effectively interpret. Healthcare providers must depend on their individual expertise and established clinical guidelines instead of systematically extracting knowledge from thousands of comparable patient cases, resulting in a significant disconnect between the potential of personalized, evidence-based medicine and actual clinical implementation. This deficiency highlights the critical necessity for more personalized, evidence-based therapeutic recommendation frameworks \citep{bhuyan2025generative, esmaeilzadeh2024challenges}. 

Individual patients exhibit distinct characteristics and demonstrate varied responses to treatments, while conventional protocols based on extensive clinical studies frequently cannot accommodate such variability, resulting in inadequate therapeutic outcomes or preventable adverse effects \citep{kumar2023artificial, xian2024language, goktas2025shaping}. Sophisticated computational methodologies, especially machine learning approaches, present a viable pathway by identifying subtle patterns and therapy-response correlations within extensive historical clinical datasets \citep{bhuyan2025generative, xian2024language}. Reinforcement learning (RL) and multi-armed bandit frameworks, particularly, are appropriately designed to enhance therapeutic approaches progressively through learning from patient results \citep{kumar2023artificial, villar2015multi}. In contrast to fixed randomized controlled trials, these dynamic systems improve decision-making by maintaining equilibrium between investigating novel therapies and utilizing established effective treatments, positioning them as optimal solutions for evolving clinical settings \citep{varatharajah2022contextual, jayaraman2024primer}.

Recent developments in medical artificial intelligence demonstrate an expanding use of RL and bandit methodologies to enhance clinical decision-making processes \citep{jayaraman2024primer}. These approaches show particular effectiveness in sequential decision scenarios, including treatment selection and dosage modification, through their capacity to integrate feedback mechanisms and customize responses based on individual patient profiles \citep{zhalechian2022online}. Multi-armed bandit frameworks effectively navigate the exploration-exploitation tension by evaluating emerging treatment modalities while giving precedence to therapies with established success rates \citep{varatharajah2022contextual}. This responsive capacity is particularly vital in learning health systems, where ethical principles require emphasizing validated therapeutic interventions while preserving continuous advancement in clinical decision support \citep{villar2015multi}.

Concurrently, Large Language Models (LLMs) including GPT-4 have revolutionized the analysis of unstructured EHR information, enabling the derivation of significant insights from clinical documentation, patient records, and discharge reports \citep{zhang2024generative, chung2025artificial}. These capabilities have been successfully applied to critical clinical prediction tasks, such as predicting hospital admissions from emergency department patient data \citep{pasquadibisceglie2025leveraging}. Through the ability to synthesize patient backgrounds and detect important clinical indicators, LLMs deliver comprehensive contextual data that can strengthen treatment recommendation systems \citep{xian2024language,xu2025staf}. Emerging studies propose integrating LLMs with bandit methodologies, leveraging LLMs to process raw textual data and produce informative features that guide personalized treatment strategies, thus minimizing early exploratory phases \citep{alamdari2024jump}.

These developments notwithstanding, substantial obstacles remain. In bandit algorithms, the cold-start challenge emerges when making initial treatment decisions for previously unseen patients or novel therapies without historical data, potentially resulting in suboptimal choices and creating safety concerns in clinical environments. Additionally, incorporating unstructured data into bandit systems presents complexity, as LLM-generated outputs may contain extraneous information that must be carefully filtered to maintain dependability in critical decision-making scenarios. The development of frameworks that effectively merge LLM-derived insights with conventional clinical variables continues to be an unresolved challenge.

Surmounting these barriers presents considerable potential benefits. An integrated framework that combines reinforcement learning algorithms with LLM-augmented clinical record interpretation would facilitate continuous improvement via patient response data, enhancing the probability of treatment efficacy. Through provision of personalized therapeutic recommendations based on individual patient narratives, this methodology could reduce futile intervention attempts while improving health outcomes such as disease stabilization and patient quality of life \citep{alamdari2024jump, xian2024language}. This framework would also streamline healthcare delivery by supporting swift, evidence-based clinical choices, especially vital in acute care environments, while minimizing superfluous diagnostics and preventable hospital returns. As an evolving medical learning platform, it would continuously integrate fresh clinical insights, including results from new therapeutic interventions, ensuring sustained improvement in treatment recommendations. This methodology could effectively unite algorithmic healthcare solutions with frontline medical practice, ensuring appropriate therapies reach suitable patients at the optimal moment.

Enabling continuous improvement through experience is accomplished by simulating the problem in online decision-making algorithms with LLMs-processed data. The proposed online personalized recommendation with continuous improvement emphasizes the importance of considering patient needs, moving beyond what traditional approaches can achieve. The remainder of the paper is structured as follows: Section~\textcolor{blue}{2} discusses the existing related literature. Section~\textcolor{blue}{3} defines the problem. Section~\textcolor{blue}{3} presents the methodology applied to the problem. Section~\textcolor{blue}{4} presents our case study. Section~\textcolor{blue}{5} reports the experimental results and evaluates the performance of the RL-based method. Section~\textcolor{blue}{6} concludes the paper and introduces directions for future work.

\section{Literature Review}

Recent advances in clinical decision science have evolved along three critical dimensions: dynamic treatment regimens, using unstructured EHRs and computational methods for making decisions. Our proposed framework sits at the intersection of these research streams, addressing fundamental gaps in each while creating a novel synthesis that promises to transform how patient history informs real-time treatment decisions and recommendation.

The design of modern clinical trials has rapidly evolved from fixed-randomization protocols to sophisticated adaptive strategies that seek both to improve patient outcomes during the trial and to yield more informative post-trial inferences. Early work in oncology and other fields highlighted the inefficiencies of one-size-fits-all trials, where average effects often mask critical heterogeneity across patients \citep{yankeelov2024designing}. In response, researchers have proposed mathematical frameworks that convert each patient into their own personalized trial, leveraging biology-based models, digital twins, optimal control theory, and data assimilation to continually refine treatment protocols \citep{yankeelov2024designing}.

This methodological evolution has brought response-adaptive randomization (RAR) to the forefront as a key technique, implemented via multi-armed bandit algorithms. \citet{norwood2024adaptive} were the first to create Thompson sampling RAR approaches for sequential multiple assignment randomized trials. Their research revealed that when randomization probabilities reflect current treatment performance beliefs, patients experience improved outcomes during the trial period while researchers maintain strong statistical power for later analyses. \citet{aslanyan2025bayesian} took a different approach by focusing on Bayesian RAR for continuous outcomes while paying special attention to covariate adjustment, and their results showed that when covariates are properly adjusted, RAR successfully assigns more patients to better treatments while keeping important baseline characteristics balanced across groups.

Acknowledging the essential requirement to identify heterogeneous treatment effects, \citet{wei2025adaptive} established a comprehensive frequentist approach for adaptive experimental designs that progressively adjust both treatment allocation probabilities and subgroup enrichment ratios. Through conceptualizing subgroup detection as an optimal-arm identification challenge within a large-deviation framework, they showed enhanced performance and diminished winner's-curse effects relative to static experimental designs across simulated e-commerce and medical applications.

Dynamic treatment regimens have been developed to overcome the limitations of static, one-size-fits-all clinical trial protocols by conceptualizing medical decision-making as an adaptive, individualized sequence of choices. Within this framework, optimal treatment is not represented by a single therapeutic option but rather by a decision-making algorithm that connects a patient's changing clinical profile to subsequent interventions, thus incorporating time-dependent variables and treatment response delays. RL constitutes an effective computational framework by casting treatment selection as a Markov Decision Process, where states characterize patient conditions, actions signify medical treatments, and rewards measure therapeutic success. Distinguished from supervised learning paradigms, RL fundamentally tackles the exploration-exploitation challenge, formulating policies that simultaneously optimize clinical outcomes and enhance the precision of future treatment choices. \citet{oh2022reinforcement} illustrated the potential of contextual bandits for diabetes care management utilizing South Korean electronic health record datasets. Their meticulously calibrated Thompson-sampling-based contextual bandit algorithm expanded the dimensionality of state and action spaces well beyond the capacity of conventional MDPs, producing individualized medication protocols that exceeded the performance of standard algorithms in retrospective analysis.

Complementing policy learning, contextual bandits have been adapted for operational challenges such as appointment scheduling and resource allocation. \citet{zhalechian2022online} introduced an algorithm combining adversarial-robust resource-allocation with Bayesian bandit learning under feedback delays. This approach strikes a three-way balance—exploration of poorly-understood decision options, exploitation of known high-reward assignments, and hedging against uncertain future demand—demonstrating superior performance in both simulated settings and a clinical appointment dataset from a major health system.

\citet{berrevoets2022treatment} confront the issue of non-stationary treatment effects by redesigning uplift modeling to prioritize individual treatment effect optimization. Their technique modifies the optimization target to boost uplift estimate precision in streaming, non-i.i.d. conditions, implementing adaptively tuned allocation probabilities to respond to changing patient-treatment response mechanisms. Empirical testing on synthetic and real datasets shows that their uplifted bandit framework successfully navigates concept drift and delivers improved performance against static conditional average treatment effect estimators in time-varying environments.

The multi-armed bandit framework has long provided a principled way to balance exploration and exploitation in sequential decision problems, and its extension to continuum-armed settings has made it directly applicable to dose-finding in early-stage clinical trials. \citet{chen2024adaptive} introduced nonparametric, continuum-armed bandit algorithms that simultaneously learn dose-efficacy and dose-toxicity curves—assumed unimodal and monotonic, respectively—without imposing rigid parametric forms. They proved regret bounds for both a dose-escalation-stop-and-commit strategy and a bisection-search-plus-upper-confidence-bound algorithm, showing substantial improvements over traditional 3+3 and equal-randomization designs. 

\citet{varatharajah2022contextual} studied personalized treatment assignment by treating trial allocations as contextual multi-armed bandits, using patient data to inform Thompson sampling or upper confidence bound algorithms that learn treatment effectiveness. When evaluated retrospectively against the International Stroke Trial dataset, their contextual bandit produced 72.6\% more participants receiving optimal therapy than standard randomization, and 64.3\% more than non-contextual bandits, showing the value of incorporating patient context into treatment decisions. The applicability of contextual bandits spans additional healthcare domains, as illustrated by \citet{parvin2019personalized}, who developed a contextual multi-armed bandit framework for behavioral anomaly identification and customized health recommendation provision in ambient-assisted living systems for elderly users, revealing the wide-ranging utility of context-informed bandit strategies in healthcare environments.

A remaining challenge for all bandit methods is the cold-start problem: how to avoid random initial assignments that waste opportunities to benefit early participants \citep{qian2025personalized}. \citet{alamdari2024jump} proposed a contextual bandits algorithm, using large language models to generate synthetic user-arm preference data. By prompting LLMs for approximated human responses, they pretrained contextual bandits that, when fine-tuned online, reduced early regret by 14-20\% across both simulated and real-world conjoint-survey settings—even under partial context obfuscation. Complementing this LLM-based approach, \citet{badreddine2019injecting} showed how to inject rich, symbolic prior knowledge directly via Logic Tensor Networks—first-order logic grounded in neural layers—augmenting pixel-based deep reinforcement learning agents with facts about object semantics and task structure to accelerate learning in grid-world tasks.

The advent of LLMs has transformed clinical information extraction, enabling zero-shot and few-shot approaches that bypass the need for extensive labeled corpora. \citet{del2025comparative} demonstrate that instruction-fine-tuned open-access models like Llama 2 and Mistral outperform base and chat-tuned variants on zero-shot clinical named entity recognition—achieving up to 71.3 F1—yet still trailing specialized supervised models by nearly 20 points. \citet{hsu2025leveraging} additionally demonstrate that LLMs can produce weak supervision signals for subsequent BERT architectures, where Llama2-13B-driven frameworks achieve performance levels that match or surpass fully supervised benchmarks even under conditions of limited gold-standard labeling. This body of research highlights the potential of LLMs for efficiently converting unstructured clinical documentation into structured formats, while simultaneously exposing their vulnerability to generating hallucinated content and requiring substantial computational resources, thereby emphasizing the critical importance of developing uncertainty estimation approaches when such model outputs inform clinical decision-making frameworks.

Retrieval-augmented generation methodologies have shown considerable promise for simultaneously improving computational efficiency and extraction performance in clinical applications. \citet{lopez2025clinical} developed a system that employs medical entity extraction to guide the retrieval mechanism, delivering a 70\% decrease in both token consumption and computational time relative to conventional full-note retrieval-augmented generation approaches, while concurrently attaining an average F1 score of 0.90. In parallel work, \citet{gu2025scalable} investigated the effectiveness of publicly available open-source LLMs for social determinants of health extraction from EHRs, reporting mention-level accuracy enhancements of up to 40\% relative to baseline pattern-matching approaches. Although these combined extraction methodologies provide marked performance improvements, they fall short of incorporating the extracted features into subsequent treatment optimization pipelines or establishing patient-specific probabilistic priors.

\citet{xie2025medical} develop Me-LLaMA—continuously pretrained on 129 billion biomedical tokens and instruction-tuned on clinical notes—outperforming both general LLaMA2 and other open medical LLMs in zero-shot and supervised settings, and rivaling GPT-4 on complex diagnosis benchmarks. \citet{akbasli2025leveraging} similarly fine-tune GPT-3 on Turkish pediatric electronic health records, achieving 99.9\% named entity recognition accuracy on respiratory infection labels and matching domain experts. These efforts illustrate how continual pretraining and instruction tuning enrich LLM knowledge, yet they rarely align extracted representations with sequential decision frameworks or quantify epistemic uncertainty when used in high-stakes contexts.

In the realm of decision support, pretrained LLMs have begun to inform predictive analytics directly from clinical records. \citet{alba2025foundational} show that self-supervised fine-tuning of ClinicalBERT and bioGPT on perioperative notes increases AUROC by up to 38\% for six postoperative risks, compared to traditional embeddings. \citet{zhang2025critical}'s technical note further highlights LLMs' capacity to predict sepsis and generate discharge summaries in critical care, streamlining workflows and flagging subtle trends in electronic health records. Despite these advances, such models predominantly focus on static risk estimation rather than on dynamic treatment recommendation, and they lack mechanisms to incorporate structured priors derived from narrative data.

Within the domain of clinical decision support, pretrained LLMs have started to directly enhance predictive analytics using clinical documentation. \citet{alba2025foundational} demonstrate that self-supervised fine-tuning of ClinicalBERT and bioGPT on perioperative documentation results in improvements of up to 38\% across six postoperative risk categories when compared to conventional embedding approaches. \citet{zhang2025critical}'s technical report additionally underscores the ability of LLMs to forecast sepsis occurrence and produce discharge summaries in critical care environments, thereby optimizing clinical workflows and identifying nuanced patterns within EHRs. However, these developments primarily concentrate on static risk assessment rather than dynamic treatment guidance, and they do not provide mechanisms for integrating structured priors extracted from narrative clinical data.RetryClaude can make mistakes. Please double-check responses.

Modern treatment effect estimation increasingly relies on counterfactual reasoning frameworks, which directly tackle the fundamental question of "what results would have emerged under different treatment choices?" rather than simply documenting patterns found in existing data \citep{wu2024clinical, prosperi2020causal}. This evolution reflects an understanding that medical decision-making necessarily involves choosing from multiple available interventions, though only the consequences of the selected intervention can be directly observed. Through the structured modeling of hypothetical outcomes across various treatment alternatives, counterfactual approaches support both improved policy analysis and the creation of decision-support technologies that can identify optimal treatment pathways for specific patients.

\citet{wu2024clinical} present a bootstrapping-based counterfactual inference method specifically constructed for observational clinical data, establishing a policy evaluation framework where therapeutic decisions constitute actions and patient medical trajectories form the contextual environment. By merging bootstrap uncertainty estimation with adversarial learning mechanisms for off-policy optimization, their technique achieves a 30\% variance reduction in counterfactual policy estimates and delivers reward improvements reaching 3\% compared to standard baseline algorithms. This study exemplifies how systematic counterfactual analysis can strengthen the credibility and safety of data-driven clinical decision-making in environments where real-time experimental validation is not viable.

Researchers have developed hybrid AI methodologies that incorporate domain expertise to improve counterfactual prediction performance. \citet{huang2024hybrid} introduce a knowledge graph-driven approach that systematically represents patient characteristics, therapeutic interventions, and clinical outcomes within an organized semantic framework, facilitating counterfactual reasoning through integrated graph-based computation and machine learning techniques. Although this method enables tailored counterfactual predictions while minimizing dependence on extensive experimental data, its performance is heavily influenced by the reliability and completeness of the knowledge graph foundation and the effectiveness of graph expansion protocols.

Graphical causal models offer another complementary path. \citet{kyrimi2025counterfactual} demonstrate how causal Bayesian networks can support counterfactual reasoning in healthcare governance contexts—such as morbidity and mortality review—by explicitly modeling treatment pathways and patient states across multiple stages of care. Their framework enables retrospective queries like "what would patient outcomes have been under an alternative treatment protocol?", but it requires careful elicitation of causal structure and probabilities from domain experts.

Despite significant progress in dynamic treatment regimens, some limitations of the current literature continue. Table~\ref{tab:lit-review} provides an overview of the literature review. While existing methods have online decision-making, and LLM-driven decision support system, they often operate independently, lacking integration of unstructured clinical data with sequential decision-making frameworks. Current dynamic treatment approaches, such as reinforcement learning or contextual bandits, typically depend on structured data and struggle with the cold-start problem, resulting in inefficient initial treatment assignments. In addition, Counterfactual models like T-learners estimate individualized treatment effects but rarely incorporate real-time, adaptive policy learning. Our research overcomes these limitations through the following key contributions:

\begin{itemize}
    \item Leveraging LLMs to extract clinically relevant features from unstructured electronic health records;
    
    \item Using these features to train T-learner models that predict treatment effects under alternative interventions;
    
    \item Employing Conditional Tabular Generative Adversarial Networks (CTGAN) to generate realistic synthetic patient data to enhance dataset robustness to tackle the cold-start problem;

    \item Initializing bandit algorithms with prior knowledge from counterfactual models for efficient learning via simulation; and

    \item  Conducting a case study focused on stage III colon cancer treatment optimization transitioning from manual assignments to digital solutions.
\end{itemize}

\begin{table}[htbp]
    \centering
    \scriptsize 
    \caption{Literature Review Table}
    \vspace{6pt}
    \label{tab:lit-review}
    \renewcommand{\arraystretch}{1.05} 
    \setlength{\tabcolsep}{5pt} 
    
    \begin{tabular}{@{}p{2.5cm}p{2.2cm}p{2.5cm}*{10}{c}p{2.5cm}@{}}
    \toprule
    \textbf{Ref.} & \textbf{Setting} & \textbf{Method. focus} &
    \rotatebox{90}{\textbf{LLM}} &
    \rotatebox{90}{\textbf{Clinical notes}} &
    \rotatebox{90}{\textbf{Counterfactual}} &
    \rotatebox{90}{\textbf{T-Learner}} &
    \rotatebox{90}{\textbf{Bandit}} &
    \rotatebox{90}{\textbf{Personalized}} &
    \rotatebox{90}{\textbf{Cold-Start Problem}} &
    \rotatebox{90}{\textbf{Prior-Informed}} &
    \rotatebox{90}{\textbf{Simulation}} &
    \rotatebox{90}{\textbf{Real-World Case}} \\
    \midrule
    \citep{norwood2024adaptive} & Clinical trials & Thompson-sampling RAR &  &  &  &  & $\checkmark$ &  &  &  & $\checkmark$ & $\checkmark$ \\
    \citep{aslanyan2025bayesian} & RCT & Covariate-adjusted Bayesian RAR &  &  &  &  & $\checkmark$ &  &  &  & $\checkmark$ & $\checkmark$ \\
    \citep{wei2025adaptive} & Clinical trials & Subgroup-adaptive best-arm ID &  &  &  &  & $\checkmark$ &  &  &  & $\checkmark$ &  \\
    \citep{chen2024adaptive} & Dose finding & Continuum non-parametric bandit &  &  &  &  & $\checkmark$ &  &  &  &  &  \\
    \citep{liberali2025real} & RCT & Real-time Gittins index &  &  &  &  & $\checkmark$ &  & $\checkmark$ &  & $\checkmark$ & $\checkmark$ \\
    \citep{varatharajah2022contextual} & Medical recommendations & Contextual Thompson-sampling bandit &  &  &  &  & $\checkmark$ & $\checkmark$ &  &  &  & $\checkmark$ \\
    \citep{oh2022reinforcement} & Medical recommendations & Contextual bandit &  & $\checkmark$ &  &  & $\checkmark$ & $\checkmark$ &  &  &  & $\checkmark$ \\
    \citep{berrevoets2022treatment} & Clinical trial & Uplifted contextual bandit &  &  & $\checkmark$ &  & $\checkmark$ & $\checkmark$ &  &  &  &  \\
    \citep{alamdari2024jump} & Recommendation systems & LLM-initialized contextual bandit & $\checkmark$ &  &  &  & $\checkmark$ & $\checkmark$ & $\checkmark$ & $\checkmark$ & $\checkmark$ &  \\
    \citep{badreddine2019injecting} & - & Transfer learning in Reinforcement Learning &  &  &  &  & $\checkmark$ & $\checkmark$ & $\checkmark$ & $\checkmark$ & $\checkmark$ &  \\
    \citep{wu2024clinical} & Treatment recommendation & Bootstrap off-policy optimization &  &  & $\checkmark$ &  & $\checkmark$ & $\checkmark$ &  &  &  & $\checkmark$ \\
    \citep{xie2025medical} & Clinical text analysis tasks & LLM pre-training & $\checkmark$ & $\checkmark$ &  &  &  &  &  &  &  &  \\
    \citep{hsu2025leveraging} & Labeling clinical notes & LLM-generated labels & $\checkmark$ & $\checkmark$ &  &  &  &  &  &  &  &  \\
    \citep{lopez2025clinical} & Clinical information extraction & Entity-guided RAG & $\checkmark$ & $\checkmark$ &  &  &  &  &  &  &  &  \\
    \citep{gu2025scalable} & Clinical information extraction & Prompt engineered LLM & $\checkmark$ & $\checkmark$ &  &  &  &  &  &  &  & $\checkmark$ \\
    \citep{del2025comparative} & Labeling entities in clinical notes & Zero-shot learning LLM & $\checkmark$ &  &  &  &  &  &  &  &  &  \\
    \textbf{This study} & Treatment recommendation & LLM + T-learner + Prior-informed contextual bandit & $\checkmark$ & $\checkmark$ & $\checkmark$ & $\checkmark$ & $\checkmark$ & $\checkmark$ & $\checkmark$ & $\checkmark$ & $\checkmark$ & $\checkmark$ \\
    \bottomrule
    \end{tabular}
\end{table}

\section{Methodology}

\begin{figure}[ht]
    \centering
    \includegraphics[ width=0.7\textwidth]{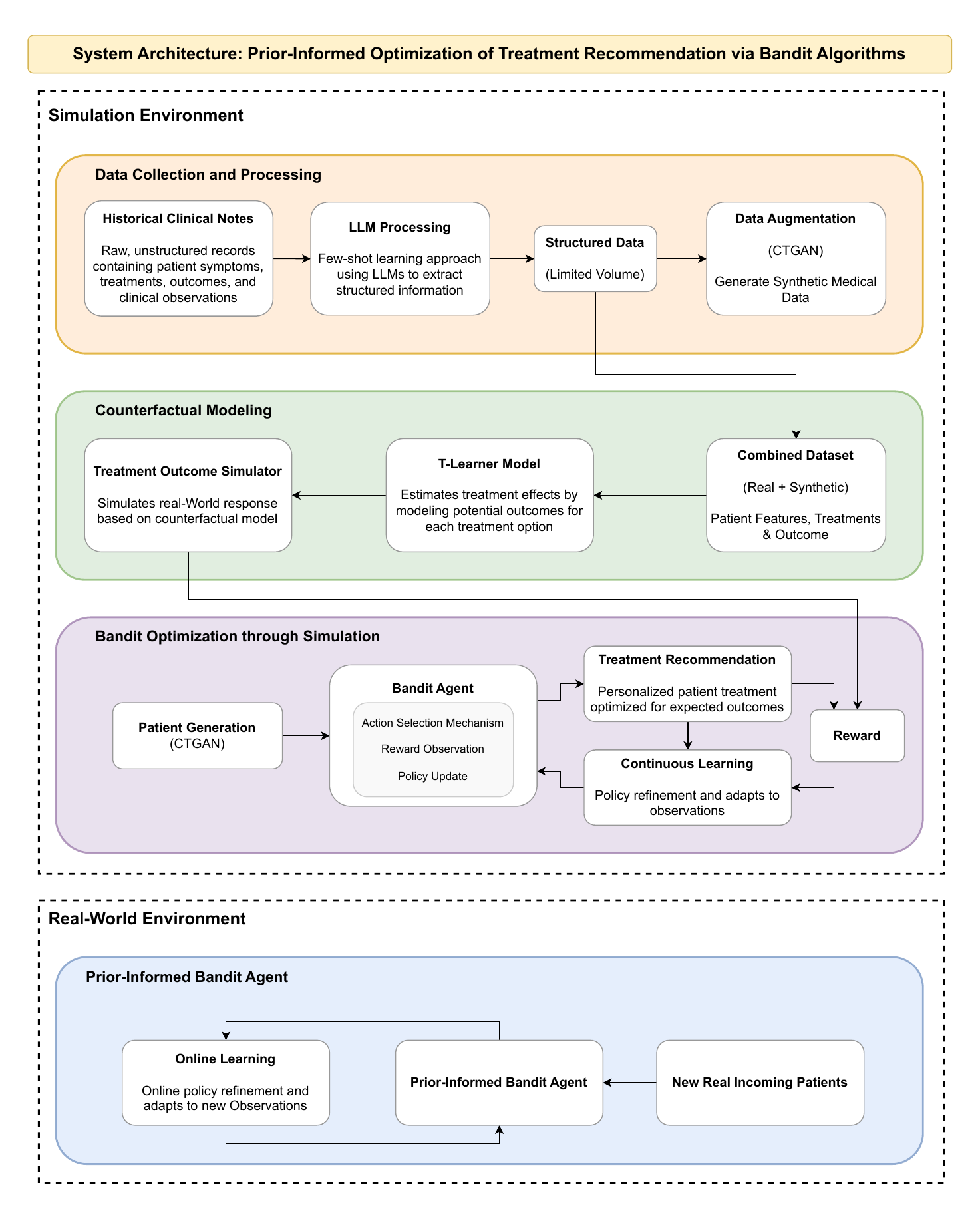}
    \caption{Treatment Recommendation System Architecture}
    \label{fig:sys-arch}
\end{figure}

Our methodological framework tackles the essential problem of delivering individualized therapeutic guidance in healthcare environments where observational data is constrained. This approach combines multiple synergistic methods to convert raw clinical documentation into practical treatment decisions while considering intervention outcomes. The overall system design is illustrated in Figure \ref{fig:sys-arch}.

We begin by processing unstructured clinical notes through a few-shot learning approach with open-source LLMs. This process transforms rich but unstructured patient information into a structured dataset suitable for algorithmic processing. To address the inherent data scarcity in clinical applications, we implement a CTGAN to generate synthetic data instances that preserve the statistical properties and relationships in the original clinical data to help generating patients for the simulation process.
With this expanded dataset, we formulate a counterfactual estimation framework employing the T-learner methodology to measure therapeutic effects across varied patient segments. This system permits the generation of predicted clinical outcomes under different treatment protocols, forming a simulation base for subsequent reinforcement learning strategies.
The ultimate methodological component integrates contextual bandit algorithms to enhance treatment selection optimization. We evaluate three differentiated computational approaches: The Linear Upper Confidence Bound (LinUCB), which postulates linear connections between context variables and outcome rewards; The Kernel Upper Confidence Bound (KernelUCB), which manages non-linear interactions via kernel-based solutions; and NeuralBandit, which applies deep learning networks to represent complex data relationships within clinical environments. Each bandit algorithm incorporates prior-informed optimization strategies that integrate domain expertise extracted from historical clinical data.
Collectively, these elements constitute an integrated pipeline that converts unstructured clinical documentation into an evidence-based treatment recommendation framework capable of maintaining equilibrium between exploring innovative therapeutic options and leveraging established effective treatments for particular patient populations. This methodology synthesizes the capabilities of natural language processing, synthetic data augmentation, causal inference, and reinforcement learning to tackle the challenges of individualized medicine within resource-limited settings.

\subsection{LLM-Processed Clinical Notes}
Our method starts by converting unstructured clinical documents into structured datasets appropriate for computational analysis. This conversion utilizes a systematic data extraction framework that employs LLMs through few-shot learning approaches. Open-source language models offer powerful textual analysis capabilities that can be configured for specialized tasks. Although these general-purpose models exhibit remarkable linguistic proficiency across diverse domains, they generally possess limited expertise in medical vocabulary, clinical decision-making processes, and healthcare record formatting standards. Instead of pursuing computationally expensive fine-tuning of these models exclusively for clinical data extraction tasks, we implemented a few-shot learning strategy that exploits the intrinsic pattern identification abilities of LLMs via strategically designed exemplars.

Few-shot learning represents a streamlined approach to domain adaptation by leveraging a minimal set of exemplary demonstrations. This strategy empowers the open-source LLM to comprehend the designated extraction procedure without requiring extensive parameter fine-tuning or specialized domain-oriented training. Through the presentation of sample medical records accompanied by their corresponding structured representations, we enable the model to detect information arrangement patterns within clinical documentation and apply these conversion processes to previously unencountered medical texts.

\begin{figure}[ht]
    \centering
    \includegraphics[ width=0.7\textwidth]{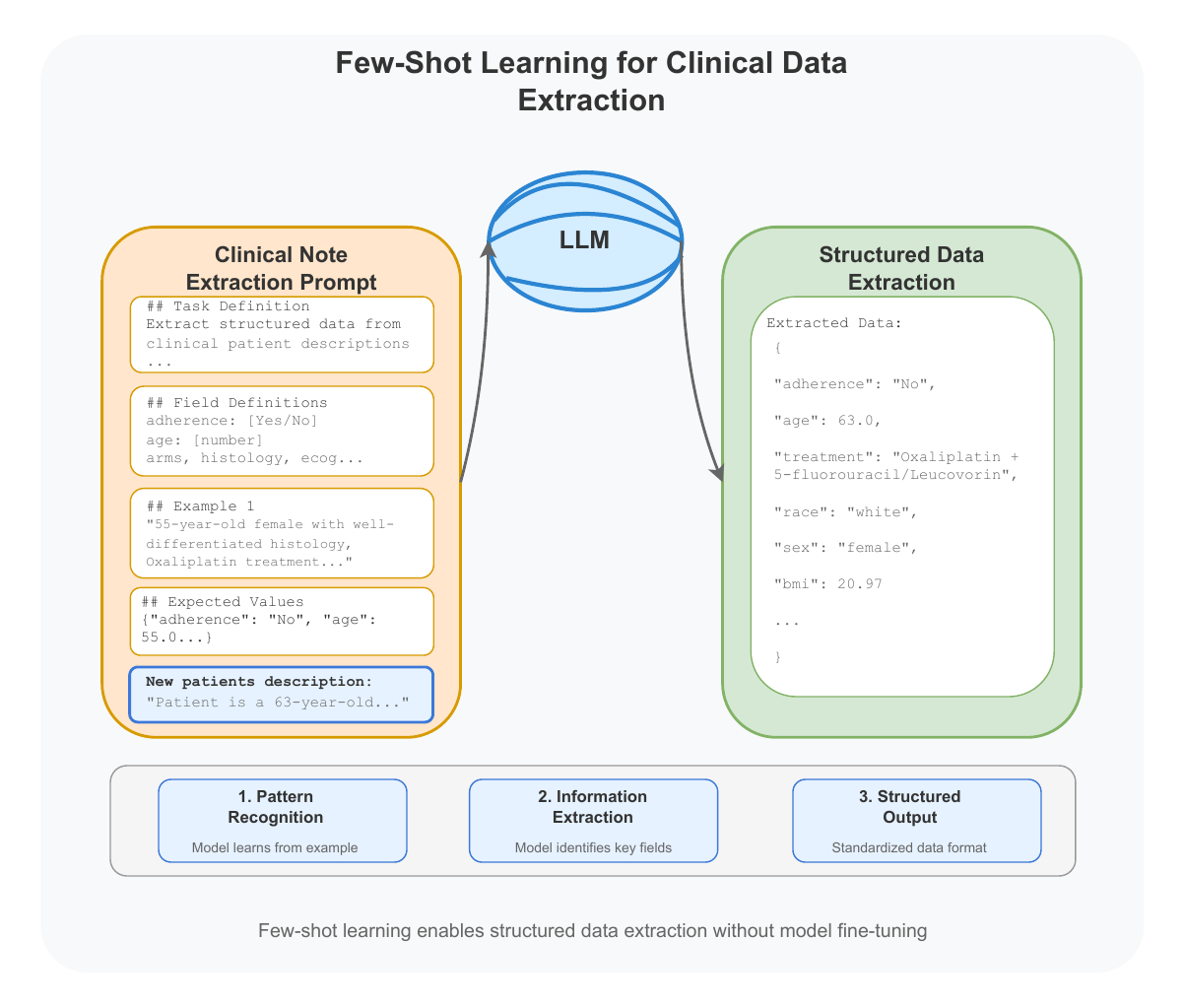}
    \caption{Clinical Few-shot Learning Process}
    \label{fig:few-shot-diag}
\end{figure}

As illustrated in Figure \ref{fig:few-shot-diag}, we initially developed a clinical data processing pipeline designed to identify and extract pertinent information from historical medical records. This process encompasses multiple interconnected stages that collaborate to transform raw clinical text into structured, usable data.

The preprocessing of clinical documentation establishes the core foundation of our pipeline system. These prompts incorporated comprehensive extraction parameters that articulated the essential data fields, supported by 3-5 thoughtfully curated representative pairs of clinical notes and their aligned structured outputs. We also embedded specialized handling mechanisms for medical abbreviations and clinical jargon particular to the healthcare sector. This few-shot approach enabled the model to develop proficiency in the extraction process without requiring intensive fine-tuning methodologies.

For the LLM implementation, we employed an open-source model architecture containing 8B parameters. The model was set up with an 8,192-token context window to handle extended clinical documentation. We configured the temperature parameter at 0.2 to encourage consistent extraction performance, and utilized greedy decoding as the sampling approach to ensure maximum reproducibility in our extraction results.

A comprehensive data extraction framework was designed with hierarchical connections between diverse clinical components. The framework encompassed patient demographic characteristics, medical backgrounds, current clinical evaluations, therapeutic protocols, and outcome assessments. This organized methodology facilitated systematic collection and arrangement of all relevant clinical information.

The extraction implementation phase entailed processing clinical documentation through the LLM pipeline using batches of 8. We employed parallel processing across multiple computational instances to enhance throughput and operational efficiency. Each document was processed using identical pipeline procedures to maintain consistency in extraction approach.

After extraction, the structured data was harmonized to maintain consistency across all records. This harmonization encompassed the standardization of measurement units, normalization of treatment terminology, temporal synchronization of treatment and outcome observations, and the resolution of conflicting or redundant information. This harmonization procedure was essential for generating a unified dataset from heterogeneous clinical documentation approaches.
This LLM-driven extraction methodology facilitated the conversion of thousands of historical clinical notes into structured data components ready for subsequent modeling processes. The final dataset constituted the basis for our treatment recommendation system, encapsulating the intricate patterns of historical treatment decisions and their corresponding outcomes.

\subsection{Conditional Tabular Generative Adversarial Network}

Following the transformation of unstructured clinical notes into structured data via our LLM-based extraction framework, we faced a prevalent issue in clinical research: limited data availability for certain treatment-outcome pairings. To overcome this constraint, we employed CTGAN to enhance our dataset with synthetically created instances that maintain the statistical characteristics of the original data.

CTGAN builds upon conventional GAN architecture by incorporating specialized modules tailored for processing tabular data containing both categorical and continuous variables. The system comprises a generator network $G$ and a discriminator network $D$ participating in an adversarial training procedure.

The generator $G$ seeks to create synthetic patient records that are indistinguishable from authentic clinical data, while the discriminator $D$ strives to distinguish between genuine and synthetic records. This adversarial procedure is expressed through the following minimax objective function:

\begin{equation}
\min_G \max_D V(D, G) = \mathbb{E}_{x \sim p_{data}(x)}[\log D(x)] + \mathbb{E}_{z \sim p_z(z)}[\log(1 - D(G(z)))]
\end{equation}

\noindent Where $p_{data}(x)$ signifies the actual clinical data distribution, $p_z(z)$ denotes a prior distribution (usually Gaussian noise), $D(x)$ reflects the discriminator's judgment of data genuineness, and $G(z)$ represents the generator's resulting output from noise input $z$.

To adequately address the diverse data types characteristic of clinical records, we implemented targeted normalization strategies for continuous variables and conditional generation methods for categorical variables. For continuous features, we deployed variational Gaussian mixture models to characterize multimodal distributions:

\begin{equation}
p(x) = \sum_{i} \pi_i \mathcal{N}(x | \mu_i, \sigma_i^2)
\end{equation}

\noindent Where $\pi_i$ reflects the weight of the $i$-th mixture component, and $\mathcal{N}(x | \mu_i, \sigma_i^2)$ denotes a Gaussian distribution with mean $\mu_i$ and variance $\sigma_i^2$.

For categorical variables, we adopted conditional vector embedding and structured the generator to deliver one-hot encoded representations. The conditional generation was established through feature-wise conditioning, where the generator develops outputs based on predetermined categorical features:

\begin{equation}
G(z, c) \rightarrow \hat{x}
\end{equation}

\noindent Where $c$ represents the conditional vector encoding specific categorical features.

The CTGAN training process involved several key modifications to enhance stability and quality of synthetic data:
\begin{itemize}
\item Gradient penalty regularization term to enforce Lipschitz continuity
\item Mode-specific normalization to handle multimodal continuous features
\item Training-by-sampling to address imbalanced categorical distributions
\end{itemize}

\begin{figure}[ht]
    \centering
    \includegraphics[ width=0.7\textwidth]{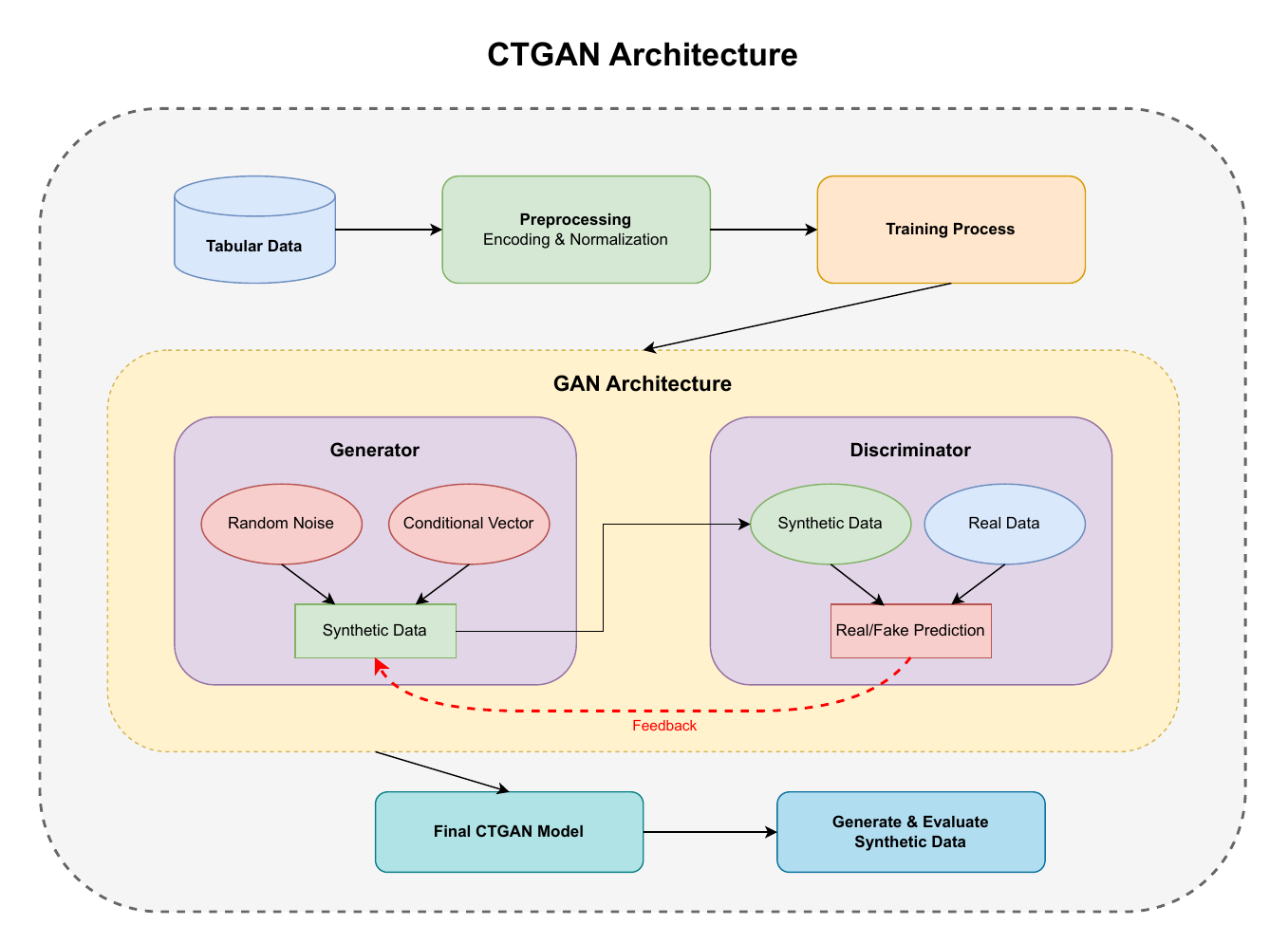}
    \caption{Workflow of the CTGAN Model}
    \label{fig:ctgan-diag}
\end{figure}

\subsection{T-learner Counterfactual Modeling}

\begin{figure}[ht]
    \centering
    \includegraphics[ width=0.7\textwidth]{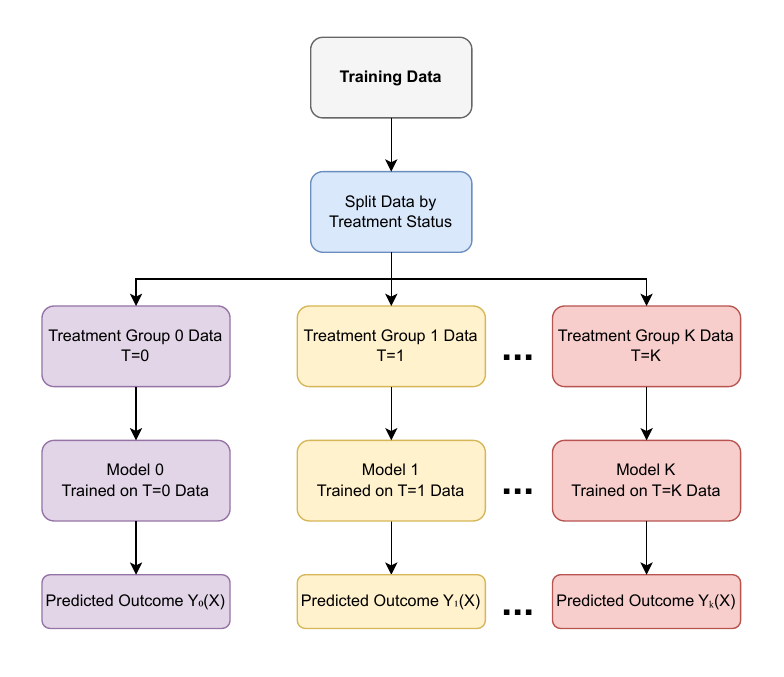}
    \caption{Structure of the T-Learner Approach}
    \label{fig:t-learner-diag}
\end{figure}

Subsequently to dataset enhancement using the CTGAN framework, we utilized a T-learner approach for counterfactual modeling to estimate individual treatment effects, as demonstrated in Figure~\ref{fig:t-learner-diag}. The T-learner constitutes a primary technique in causal inference that enables us to calculate the potential outcomes under different treatment regimens for each patient.

The T-learner framework necessitates constructing separate prediction models for each treatment condition, allowing us to deduce what would have occurred under alternative treatment scenarios. This approach is especially beneficial for our clinical recommendation system as it captures complex, non-linear interactions between patient characteristics and treatment effectiveness.

For our implementation, we denote the potential outcome for a patient with features $X$ under treatment $t$ as $Y(t)$. With multiple possible treatments $T = {t_1, t_2, \ldots, t_k}$, the T-learner trains $k$ separate models $\mu_t$, each estimating $\mathbb{E}[Y|X, T=t]$ for a specific treatment $t$.
The formal representation of our T-learner framework can be expressed as:

\begin{equation}
\mu_t(x) = \mathbb{E}[Y|X=x, T=t]
\end{equation}

For each treatment option $t$, we train a dedicated model $\mu_t$ using only the subset of data where treatment $t$ was administered. The models are trained to minimize the empirical risk:

\begin{equation}
\min_{\mu_t} \sum_{i:T_i=t} L(\mu_t(X_i), Y_i)
\end{equation}

\noindent Where $L$ represents the loss function, which in our case was mean squared error for continuous outcomes and cross-entropy loss for categorical outcomes.
For each patient with features $x$, the predicted outcome under each treatment $t$ is given by $\mu_t(x)$. This provides us with a complete set of counterfactual outcomes:

\begin{equation}
\text{Predicted Outcomes}(x) = {\mu_{t_1}(x), \mu_{t_2}(x), \ldots, \mu_{t_k}(x)}
\end{equation}

These individual treatment outcome predictions allow us to estimate what would happen to a specific patient under each available treatment option, providing the foundation for treatment selection and recommendation. To implement the T-learner, we developed a model for each treatment arm using four different machine learning algorithms: XGBoost, Random Forest, Neural Networks, and Support Vector Machines (SVM). This diverse approach helped capture both structured relationships and complex interactions in the data

To address potential selection bias in the observational data, we implemented inverse probability of treatment weighting (IPTW) during model training. The propensity score $e_t(x)$ for receiving treatment $t$ was estimated using a separate gradient-boosted classifier:

\begin{equation}
e_t(x) = P(T=t|X=x)
\end{equation}

The inverse probability weights were then calculated as:

\begin{equation}
w_i = \frac{\mathbb{I}(T_i = t)}{e_t(X_i)}
\end{equation}

\noindent Where $\mathbb{I}(T_i = t)$ is an indicator function that equals 1 when patient $i$ received treatment $t$ and 0 otherwise.

These weights effectively create statistical copies of observations, generating a pseudo-population where the covariate distributions are balanced across treatment groups. This reweighting procedure increases the sample size while ensuring that treatment assignment is no longer confounded by patient characteristics.

These weights were incorporated into the loss function during T-learner training to mitigate selection bias:

\begin{equation}
\min_{\mu_t} \sum_{i:T_i=t} w_i \cdot L(\mu_t(X_i), Y_i)
\end{equation}

The trained T-learner models provided us with a robust mechanism to estimate counterfactual outcomes for each treatment arm, effectively simulating what would happen to a patient under each potential treatment option. These individual treatment outcome estimates formed the foundation for our subsequent bandit algorithms, enabling them to learn from simulated feedback in scenarios where real-world experimentation would be impractical or unethical.

\subsection{Contextual Bandit Framework}

After constructing our counterfactual model, we adopted a contextual bandit framework to improve treatment recommendations. Contextual bandits present a sophisticated approach to the exploration-exploitation challenge inherent in sequential decision-making tasks, particularly appropriate for our clinical recommendation setting where each patient constitutes a unique situation demanding tailored treatment selection.

The contextual bandit framework operates on the principle of learning optimal policies through environmental interaction. In our implementation, the bandit agent examines patient features (context), chooses a treatment (action), and receives feedback on the treatment's performance (reward). This approach can be represented as follows:

At each time step $t$:
\begin{enumerate}
\item The agent observes a context $x_t \in \mathcal{X}$ representing a patient's clinical features
\item Based on context $x_t$, the agent selects an action $a_t \in \mathcal{A}$ from the set of available treatments
\item The environment (simulated by our T-learner counterfactual model) returns a reward $r_t$
\item The agent updates its policy based on the observed triplet $(x_t, a_t, r_t)$
\end{enumerate}

The objective of the contextual bandit is to maximize the cumulative reward over time, expressed mathematically as:

\begin{equation}
\max \mathbb{E}\left[\sum_{t=1}^T r_t\right]
\end{equation}

The expected reward for a given context-action pair is defined as:

\begin{equation}
\mu(x, a) = \mathbb{E}[r|x, a]
\end{equation}

And the optimal policy $\pi^*$ selects actions that maximize this expected reward:

\begin{equation}
\pi^*(x) = \mathop{\arg\max}_{a\in\mathcal{A}} \mu(x, a)
\end{equation}

To efficiently manage the exploration-exploitation trade-off, we deployed multiple bandit algorithms incorporating prior-informed adaptations. These algorithms compute the anticipated reward for each treatment alternative while preserving uncertainty estimates to direct exploration. Below, we outline our implementations of LinUCB, KernelUCB, and NeuralBandit algorithms.

\subsubsection{LinUCB}

The LinUCB algorithm assumes a linear relationship between the context-action features and the expected reward. For our clinical recommendation system, we implemented LinUCB with prior-informed modifications to incorporate domain knowledge, as shown in Algorithm~\ref{alg:lincub}.

In LinUCB, the expected reward is modeled as:

\begin{equation}
\mu(x, a) = \theta_a^T \phi(x, a)
\end{equation}

\noindent Where $\theta_a$ represents the unknown parameter vector for action $a$, and $\phi(x, a)$ is a feature mapping of the context-action pair.

For each action $a$, the algorithm maintains a parameter estimate $\hat{\theta}_a$ and a confidence matrix $A_a$. The upper confidence bound is computed as:

\begin{equation}
\text{UCB}(x, a) = \hat{\theta}_a^T \phi(x, a) + \alpha\sqrt{\phi(x, a)^T A_a^{-1} \phi(x, a)}
\end{equation}

\noindent Where $\alpha$ is an exploration parameter controlling the width of the confidence interval.

The algorithm selects the action with the highest upper confidence bound:

\begin{equation}
a_t = \mathop{\arg\max}_{a\in\mathcal{A}} \text{UCB}(x_t, a)
\end{equation}

After observing reward $r_t$, the parameter estimates are updated using ridge regression:

\begin{align}
A_a &\leftarrow A_a + \phi(x_t, a_t)\phi(x_t, a_t)^T\\
b_a &\leftarrow b_a + r_t\cdot\phi(x_t, a_t)\\
\hat{\theta}_a &\leftarrow A_a^{-1}b_a
\end{align}

To incorporate prior knowledge derived from historical clinical data, we modified the initial parameter estimates and confidence matrices:

\begin{align}
A_a &= \lambda_a\cdot I + \sum_{i\in D_{prior}} \phi(x_i, a)\phi(x_i, a)^T\\
b_a &= \sum_{i\in D_{prior}} r_i\cdot\phi(x_i, a)
\end{align}

\noindent Where $D_{prior}$ represents a carefully curated subset of historical data points, and $\lambda_a$ is a prior-specific regularization parameter.

\begin{algorithm}
    \caption{LinUCB Algorithm}
    \label{alg:lincub}
    \begin{algorithmic}[1]
    \Require Regularization parameter $\lambda > 0$, exploration parameter $\alpha > 0$, dimension $d$, time horizon $T$
    \State Initialize $\mathbf{A}_0 = \lambda \mathbf{I}_d$ \Comment{$d \times d$ regularization matrix}
    \State Initialize $\mathbf{b}_0 = \mathbf{0}_d$ \Comment{$d$-dimensional zero vector}
    \For{$t = 1, 2, \ldots, T$}
        \State Observe context features $\mathbf{x}_{t,a} \in \mathbb{R}^d$ for all arms $a \in \mathcal{A}_t$
        \State Compute $\hat{\boldsymbol{\theta}}_{t-1} = \mathbf{A}_{t-1}^{-1}\mathbf{b}_{t-1}$ \Comment{Ridge regression estimate}
        \For{each arm $a \in \mathcal{A}_t$}
            \State $p_{t,a} = \mathbf{x}_{t,a}^{\top}\hat{\boldsymbol{\theta}}_{t-1} + \alpha\sqrt{\mathbf{x}_{t,a}^{\top}\mathbf{A}_{t-1}^{-1}\mathbf{x}_{t,a}}$ \Comment{UCB score}
        \EndFor
        \State Select arm $a_t = \arg\max_{a \in \mathcal{A}_t} p_{t,a}$
        \State Observe reward $r_t$
        \State $\mathbf{A}_t = \mathbf{A}_{t-1} + \mathbf{x}_{t,a_t}\mathbf{x}_{t,a_t}^{\top}$ \Comment{Update precision matrix}
        \State $\mathbf{b}_t = \mathbf{b}_{t-1} + r_t\mathbf{x}_{t,a_t}$ \Comment{Update weighted context sum}
    \EndFor
    \end{algorithmic}
\end{algorithm}

\subsubsection{KernelUCB}

The KernelUCB algorithm expands upon LinUCB by incorporating non-linear relationships through kernel methods, as described in Algorithm~\ref{alg:kernelbandit}. This methodology is especially beneficial for modeling complex interactions between patient characteristics and treatment responses.
KernelUCB employs a kernel function $K(\cdot,\cdot)$ to implicitly transform context-action pairs into a high-dimensional feature space. The expected reward is represented as:

\begin{equation}
\mu(x, a) = f_a(\phi(x, a))
\end{equation}

\noindent Where $f_a$ belongs to a reproducing kernel Hilbert space defined by kernel $K$.

The upper confidence bound is computed as:

\begin{equation}
\text{UCB}(x, a) = k_t(x, a)^T (K_t + \lambda I)^{-1} y_t + \beta_t\sqrt{k(x, a, x, a) - k_t(x, a)^T (K_t + \lambda I)^{-1} k_t(x, a)}
\end{equation}

\noindent Where:
\begin{itemize}
\item $k_t(x, a) = [K((x_1, a_1), (x, a)), ..., K((x_{t-1}, a_{t-1}), (x, a))]^T$
\item $K_t$ is the kernel matrix with entries $K_t[i,j] = K((x_i, a_i), (x_j, a_j))$
\item $y_t = [r_1, ..., r_{t-1}]^T$
\item $\beta_t$ is the exploration parameter at time $t$
\end{itemize}

We implemented KernelUCB with a Gaussian kernel function:

\begin{equation}
K((x, a), (x', a')) = I(a = a') \cdot \exp(-\gamma\|x - x'\|^2)
\end{equation}

\noindent Where $\gamma$ is the kernel bandwidth parameter and $I(\cdot)$ is an indicator function ensuring that kernel computations are only performed between identical actions.

To incorporate prior knowledge, we initialized the kernel matrix $K_0$ and reward vector $y_0$ using historical data:

\begin{align}
K_0[i,j] &= K((x_i, a_i), (x_j, a_j)) \text{ for all } i,j \in D_{prior}\\
y_0 &= [r_i]_{i\in D_{prior}}
\end{align}

\begin{algorithm}
    \caption{Kernel Bandit Algorithm}
    \label{alg:kernelbandit}
    \begin{algorithmic}[1]
    \Require Kernel function $k(\cdot,\cdot)$, exploration parameter $\beta > 0$, action set $\mathcal{A}$, time horizon $T$
    \State Initialize $\mathcal{D}_0 = \emptyset$ \Comment{Empty dataset}
    \For{$t = 1, 2, \ldots, T$}
        \State Compute posterior mean and variance using kernel regression:
        \State $\mu_{t-1}(a) = \mathbf{k}_t(a)^T(\mathbf{K}_t + \lambda \mathbf{I})^{-1}\mathbf{y}_t$ for all $a \in \mathcal{A}$
        \State $\sigma^2_{t-1}(a) = k(a,a) - \mathbf{k}_t(a)^T(\mathbf{K}_t + \lambda \mathbf{I})^{-1}\mathbf{k}_t(a)$ for all $a \in \mathcal{A}$
        \State Select action $a_t = \arg\max_{a \in \mathcal{A}} \mu_{t-1}(a) + \beta\sigma_{t-1}(a)$ \Comment{Upper Confidence Bound}
        \State Observe reward $r_t$
        \State Update dataset $\mathcal{D}_t = \mathcal{D}_{t-1} \cup \{(a_t, r_t)\}$
        \State Update kernel matrix $\mathbf{K}_t$ and vector $\mathbf{k}_t(a)$ for all $a \in \mathcal{A}$
    \EndFor
    \end{algorithmic}
    \end{algorithm}

\subsubsection{NeuralBandit}

The NeuralBandit framework exploits deep neural networks to represent the complex interdependencies between patient contexts, treatments, and outcomes, as depicted in Algorithm~\ref{alg:neuralbandit}. This technique provides remarkable flexibility in detecting highly non-linear structures within clinical data.

Our NeuralBandit implementation utilizes a neural network architecture to determine $Q(x, a)$, the estimated reward for context-action pairs. The network consists of shared representation layers followed by action-specific output heads:

\begin{equation}
Q(x, a) = f_a(g(x))
\end{equation}

\noindent Where $g(x)$ represents shared feature extraction layers and $f_a$ represents the action-specific output head for action $a$.

To balance exploration and exploitation, we implemented a Bayesian neural network approach using dropout as a Bayesian approximation. During decision-making, we performed multiple forward passes with dropout enabled to obtain a distribution of reward estimates:

\begin{equation}
\{Q^1(x, a), Q^2(x, a), ..., Q^M(x, a)\}
\end{equation}

The upper confidence bound was then computed as:

\begin{equation}
\text{UCB}(x, a) = \text{mean}(\{Q^j(x, a)\}_{j=1}^M) + \alpha\cdot\text{std}(\{Q^j(x, a)\}_{j=1}^M)
\end{equation}

\noindent Where $\text{mean}(\cdot)$ and $\text{std}(\cdot)$ represent the mean and standard deviation of the dropout samples, and $\alpha$ is an exploration parameter.

The NeuralBandit architecture consisted of three shared hidden layers (128, 64, 32 neurons) with ReLU activations and batch normalization, followed by treatment-specific heads with two hidden layers (16, 8 neurons).

To incorporate prior knowledge, we pre-trained the neural network on historical data before initiating the online learning process:

\begin{equation}
\min_\theta \sum_{(x,a,r)\in D_{prior}} (Q_\theta(x, a) - r)^2 + \lambda\|\theta\|^2
\end{equation}

\noindent Where $Q_\theta$ represents the neural network with parameters $\theta$, and $\lambda$ is a regularization parameter.

This Prior-Informed Optimization approach across all three bandit algorithms enabled effective leveraging of historical clinical knowledge while maintaining the adaptability needed for personalized treatment recommendations.

\begin{algorithm}
    \caption{Neural Bandit Algorithm}
    \label{alg:neuralbandit}
    \begin{algorithmic}[1]
    \Require Neural network $f_\theta$, exploration parameter $\alpha > 0$, action set $\mathcal{A}$, time horizon $T$, batch size $B$, training epochs $E$
    \State Initialize neural network parameters $\theta$ randomly
    \State Initialize replay buffer $\mathcal{D} = \emptyset$
    \For{$t = 1, 2, \ldots, T$}
        \State Generate $M$ Monte Carlo dropout samples $\{\hat{f}_{\theta_m}(a)\}_{m=1}^M$ for all $a \in \mathcal{A}$
        \State Compute mean: $\mu_t(a) = \frac{1}{M}\sum_{m=1}^M \hat{f}_{\theta_m}(a)$ for all $a \in \mathcal{A}$
        \State Compute variance: $\sigma^2_t(a) = \frac{1}{M}\sum_{m=1}^M (\hat{f}_{\theta_m}(a) - \mu_t(a))^2$ for all $a \in \mathcal{A}$
        \State Select action $a_t = \arg\max_{a \in \mathcal{A}} \mu_t(a) + \alpha\sigma_t(a)$ \Comment{Neural UCB}
        \State Observe reward $r_t$
        \State Add $(a_t, r_t)$ to replay buffer $\mathcal{D}$
        \If{$t \mod B == 0$} \Comment{Train network every $B$ steps}
            \For{$e = 1, 2, \ldots, E$}
                \State Sample mini-batch from $\mathcal{D}$
                \State Update $\theta$ by minimizing MSE loss on mini-batch
            \EndFor
        \EndIf
    \EndFor
    \end{algorithmic}
\end{algorithm}
\section{Real-World Case Study}

This case study investigates the enhancement of treatment methodologies for stage III colon cancer patients within healthcare institutions that are transitioning from traditional treatment decision-making practices to more structured systems. Presently, the healthcare institution employs conventional clinician-centered treatment selection mechanisms.. Medical oncologists manage therapeutic decisions through patient case evaluations during multidisciplinary team discussions, typically relying on clinical judgment and standardized treatment protocols without incorporating extensive data synthesis or predictive analytical tools. This approach, although maintaining clinical validity, demands considerable resources, remains vulnerable to individual practitioner biases, and becomes progressively more challenging to refine as therapeutic options continue to expand. Furthermore, the healthcare institution encounters variations in treatment outcomes, especially in complex cases where several therapeutic pathways are viable, resulting in inefficient resource allocation and variable patient outcomes.

To address these challenges, healthcare facilities can implement a more sophisticated and expandable methodology for treatment selection. Instead of depending exclusively on clinical judgment and uniform protocols, medical practices can integrate an AI-powered treatment advisory system into their operational processes. This system is engineered to provide support for critical treatment selection decisions, including determining optimal therapeutic regimens, forecasting patient responses according to individual characteristics, refining its decision-making algorithms, and choosing the most appropriate treatment protocols. Through analysis of past treatment data and patient outcomes, the AI framework can facilitate more accurate, data-driven treatment choices compared to conventional decision-making approaches. This transition is anticipated to enhance therapeutic effectiveness, minimize adverse reactions, and provide better alignment with patient-specific requirements and preferences

\begin{figure}[htbp]
\centering
\includegraphics[width=0.85\linewidth]{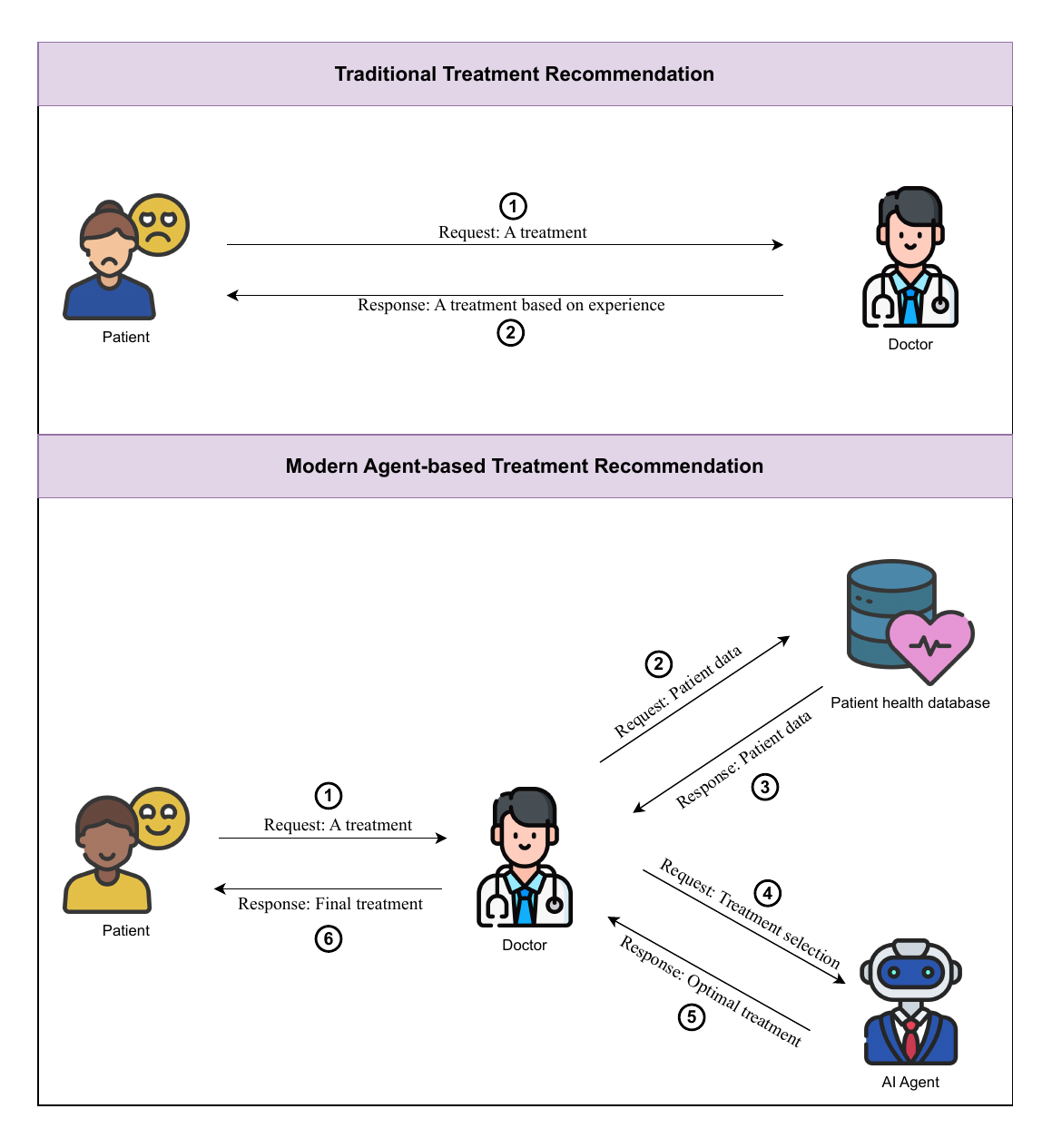}
\caption{Comparison workflow of traditional and modern treatment recommendation}
\label{fig:case_study}
\end{figure}

As part of this transition, clinics can also aim to modernize the treatment decision-making process (as shown in Figure~\ref{fig:case_study}). It is developing a system that allows clinicians to input comprehensive patient data and receive real-time treatment recommendations tailored to each patient's unique clinical profile. This digital interface will enhance clinical decision-making capabilities, increase treatment precision, and ensure that therapeutic choices account for both clinical indicators and patient-specific factors from the outset.

This case study emphasized post-surgical adjuvant chemotherapy guidance for patients after successful curative resection procedures. The recommendation challenge encompassed choosing from six therapeutic options: ARM A (oxaliplatin combined with fluorouracil/leucovorin), ARM B (irinotecan hydrochloride combined with fluorouracil/leucovorin), ARM C (sequential ARM A then ARM B), ARM D (ARM A with cetuximab), ARM E (ARM B with cetuximab), and ARM F (ARM C with cetuximab), with ARM G (locally directed therapy) subsequently incorporated for patients harboring mutant KRAS. The framework was designed to identify the most suitable treatment protocol based on individual patient parameters such as age, tumor features, lymph node status, and biomarker characteristics, thereby illustrating its capacity to individualize oncological treatment decisions within a clinically applicable context.

\section{Result}
\subsection{Clinical Note Structuring Performance}

The accurate transformation of unstructured clinical notes into structured tabular data is foundational to our treatment recommendation framework. We evaluated two open-source LLMs for this task: Llama 3.1 (8B parameters) and DeepSeek-R1 (8B parameters). Both models were assessed on their ability to extract relevant clinical features through few-shot learning approaches using an identical prompt structure with 3-5 examples.

The evaluation employed a reference standard dataset consisting of 25 manually curated clinical documents. Assessment was centered on accuracy as the main evaluation criterion, operationalized as the ratio of correctly identified fields to total fields. This metric selection corresponds to the dichotomous nature of the extraction task—where each field represents a correct or incorrect identification—thus providing accuracy as an immediate measure of the comprehensive reliability of the structured data extraction process.

Llama 3.1 attained an accuracy of 82.7\%, showing competent performance but encountering difficulties with implicit clinical context and ambiguous medical language, which resulted in extraction inaccuracies and omissions across certain fields.
In contrast, DeepSeek-R1 exhibited superior performance with 93.2\% accuracy, indicating enhanced capability in correctly identifying and excluding relevant fields relative to Llama 3.1. The model's advanced reasoning functionality facilitated improved comprehension of specialized medical patterns and complex clinical associations. Nevertheless, DeepSeek-R1 experienced occasional extraction failures, particularly when processing heavily condensed clinical abbreviations.

Regarding computational efficiency, Llama 3.1 processed clinical notes approximately 35\% faster on average, with a mean processing time of 28 seconds per note compared to DeepSeek-R1's 43 seconds when evaluated on identical hardware. This difference in processing time can be attributed to DeepSeek-R1's more complex reasoning architecture, which, while contributing to higher accuracy, inherently requires more computational resources per inference step. This efficiency difference, while modest, becomes significant when scaling to large clinical datasets.

Both models produced structured data of sufficient quality for downstream tasks. DeepSeek-R1 was selected for subsequent phases due to its higher accuracy in extracting the clinical information, ensuring a more reliable structured dataset for our treatment optimization pipeline.

These results validate the feasibility of using open-source LLMs with few-shot learning for clinical information extraction without requiring domain-specific fine-tuning, which forms the critical first step in our treatment optimization framework.

\subsection{Synthetic Data Generation and Validation}

After successfully structuring the clinical notes, we tackled the issue of limited data availability by generating synthetic data using a Conditional Tabular Generative Adversarial Network (CTGAN). This section outlines the outcomes of our synthetic data generation approach and the validation methods that followed.

\subsubsection{CTGAN Training Performance}

\begin{figure}[htbp]
    \centering
    \includegraphics[width=0.8\textwidth]{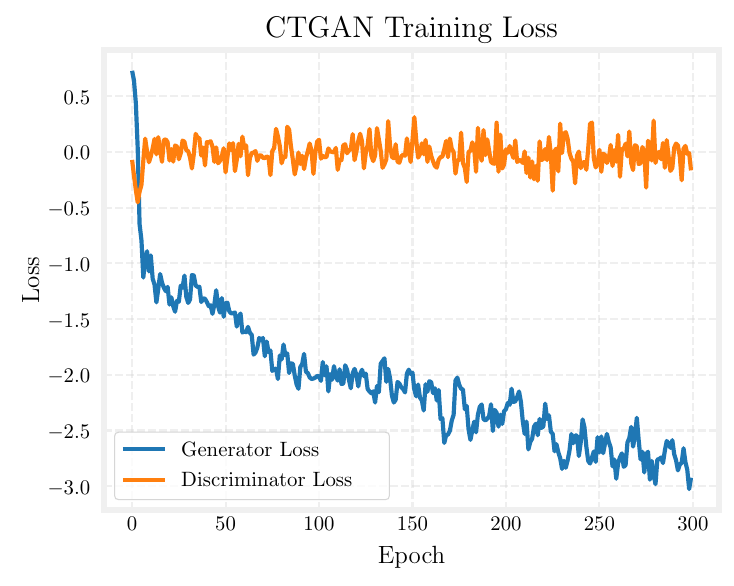}
    \caption{CTGAN training loss}
    \label{fig:ctgan-loss}
\end{figure}

Our LLM pipeline's extracted structured clinical data served as the training dataset for the CTGAN model. The training loss progression across epochs is depicted in Figure \ref{fig:ctgan-loss}. The generator loss, shown by the blue curve, initiates at a positive value before rapidly transitioning to negative values, reflecting the generator's enhanced capacity to synthesize realistic data. This downward trend in generator loss persists steadily until stabilizing at approximately -3, signifying the generator's continuous advancement in producing data capable of effectively fooling the discriminator.

The orange curve depicts the discriminator (critic) loss, which oscillates around zero throughout the training process. This pattern is characteristic of the Wasserstein GAN with gradient penalty (WGAN-GP) framework employed in our CTGAN implementation. The discriminator's relatively stable loss pattern, with minor oscillations, indicates that it maintains a balanced ability to distinguish between real and synthetic data while engaging in the adversarial training process.

The dynamics between these loss curves offer significant insights into the model's training behavior. In analyzing these adversarial loss patterns, three distinct phases emerge: In the initial phase, the generator creates suboptimal synthetic data while the discriminator experiences difficulty in distinguishing between authentic and generated samples. During the progression of training, the generator's progressively more negative loss values indicate its enhanced performance, while the discriminator must continually improve its detection capabilities to sustain steady performance levels. In the final phase, once the generator loss reaches stability at a negative value with the discriminator loss maintaining proximity to zero, it can be determined that the generator has successfully optimized its capacity to generate synthetic data that closely mirrors the authentic clinical records.

The stabilization of both loss curves following approximately epoch 100 indicates that the model achieved effective convergence at this juncture, with negligible additional enhancement in later epochs. This training behavior demonstrates that the CTGAN model effectively maintained equilibrium in the adversarial dynamics between generator and discriminator, producing a robust model with the capability to generate data that is statistically comparable to our original clinical dataset.

\subsubsection{Validation of Synthetic Data}
To rigorously evaluate the quality of the synthetic data, we employed a machine learning-based two-sample test. This approach involved labeling real clinical data instances as class 1 and synthetic data instances as class 0, then training a Random Forest classifier to distinguish between these two classes. In an ideal scenario, a high-quality synthetic dataset would be indistinguishable from real data, resulting in classification performance near random chance.

\begin{figure}[htbp]
    \centering
    \includegraphics[width=0.8\textwidth]{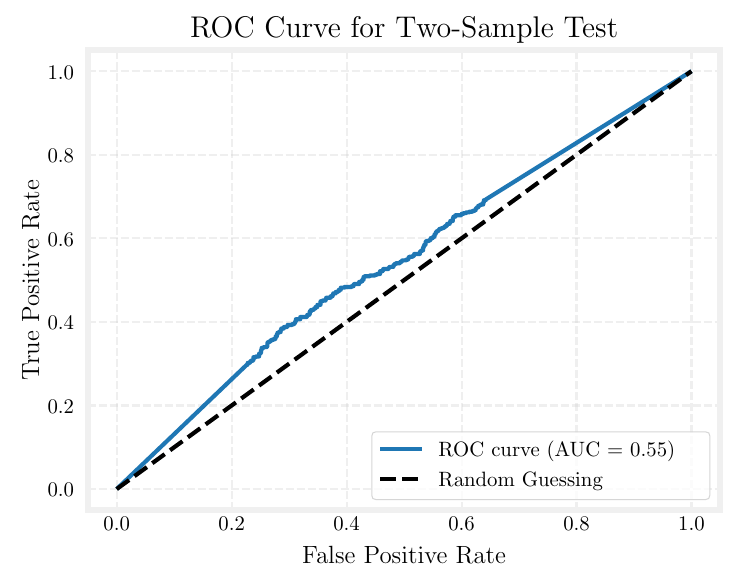}
    \caption{ROC curve for two sample test}
    \label{fig:roc}
\end{figure}

Figure \ref{fig:roc} presents the receiver-operating-characteristic (ROC) curve obtained from this classifier. The curve plots the true-positive rate (proportion of synthetic rows correctly labeled as synthetic) against the false-positive rate (proportion of real rows incorrectly labeled as synthetic) as the decision threshold is swept from strict to lenient. The dashed 45° line represents random guessing; a model that cannot exploit any systematic differences between the two samples would trace this diagonal exactly.

The area under the ROC curve stands at 0.55, which represents only a slight elevation above the random chance threshold of 0.50. This demonstrates that the classifier possesses merely a 55\% likelihood of correctly ranking a randomly selected synthetic sample higher than a randomly selected authentic sample—scarcely superior to random guessing. Put differently, the ensemble model cannot identify robust, reliable patterns that distinguish between the two distributions. This outcome implies that the synthetic generator replicates the joint distribution structure of the authentic data with substantial accuracy; any remaining discrepancies are minor and may result from limited sample size effects or slight distributional shifts rather than fundamental deficiencies in the generation methodology.
The horizontal axis of the ROC curve depicts the False Positive Rate (FPR), representing the proportion of authentic data misclassified as synthetic. The vertical axis shows the True Positive Rate (TPR), indicating the proportion of synthetic data accurately identified as synthetic. The blue curve emerges from plotting these two metrics across different decision thresholds. These thresholds constitute the decision boundaries employed by the random forest classifier to distinguish data as either authentic or synthetic. The classifier assigns each data point a probability score reflecting its likelihood of being synthetic; when this score surpasses the threshold value, the point receives a synthetic classification. As the threshold decreases, additional data points are labeled as synthetic, leading to simultaneous increases in both FPR and TPR. This produces a step-like curve that originates at the coordinate (0,0) and advances toward the endpoint (1,1).

The stepwise pattern of the curve results from the discrete threshold values and the classifier's behavior at these decision points. With the score distributions of real and synthetic data being nearly superimposed here, the curve tracks closely along the diagonal with minimal variations. These small departures indicate the subtle patterns the model identifies while trying to differentiate between the two data categories. The results demonstrate that the synthetic data maintains substantial similarity to the real data, representing a favorable outcome for synthetic data generation approaches. Nevertheless, the classification performance indicates potential for enhancement in achieving complete indistinguishability between synthetic and real data, especially in regions where the model exhibits modest yet systematic discrimination.

\subsubsection{Dimensional Analysis and Visualization}

\begin{figure}[htbp]
    \centering
    \includegraphics[width=0.8\textwidth]{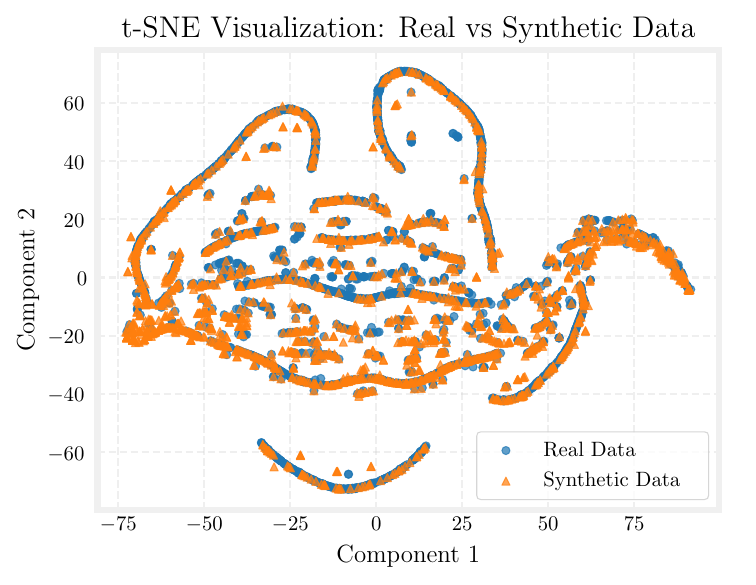}
    \caption{t-SNE visualization for patient data}
    \label{fig:tsne}
\end{figure}

The t-distributed Stochastic Neighbor Embedding (t-SNE) analysis shown in Figure \ref{fig:tsne} provides a two-dimensional visualization that contrasts the distributions of original and synthetic datasets. This dimensionality reduction method maintains the local structure of high-dimensional data by keeping neighboring points in close proximity within the reduced space while separating distant observations. The visualization employs blue circles to denote authentic data points and orange triangles to indicate synthetically generated samples.

The analysis demonstrates considerable overlap between authentic and generated data points, exhibiting similar clustering behavior and density distributions across the manifold structure. This color intermixing suggests that the CTGAN has effectively learned the complex, multidimensional relationships that characterize distinct patient populations. Critical clinical cohorts, especially those distinguished by particular therapeutic sequences and comorbidity profiles, maintain consistent grouping behaviors in both datasets. In regions where concentrated bands, curves, or clusters of original data emerge, generated samples reliably coincide with these patterns, evidencing the model's capacity to accurately represent the underlying data distribution.

A limited number of micro-regions exhibit noticeable chromatic imbalances—brief extensions or boundary portions where orange predominates or blue becomes sparse—indicating modest distributional shifts rather than substantial model deficiencies. These minor discrepancies underscore regions where the model may have either overlooked uncommon patient characteristics or excessively reproduced certain features. Notwithstanding these slight variations, the general assessment supports that an observer reviewing this visualization without access to the legend would find it challenging to distinguish between the two datasets visually, reinforcing the approximately random AUC documented in our random-forest classifier evaluation.

The t-SNE plot has several important uses in our validation work. It gives us a visual method to check how similar the synthetic and real data distributions are, backing up our statistical results. The plot also lets us find places where the original dataset has gaps that synthetic data points might cover, demonstrating that the CTGAN can produce reasonable data even in areas where we have limited real examples from the clinical feature space.

It should be emphasized that t-SNE represents a qualitative approach and exhibits sensitivity to hyperparameter settings (perplexity, learning rate, random seed). Although it accurately maintains local structural relationships, it may alter global geometric properties, meaning that overlapping distributions constitute a necessary but insufficient condition for establishing distributional equivalence. However, when considered alongside our ROC analysis, this visual convergence offers strong support that our synthetic data preserves the essential statistical characteristics of the authentic cohort while generating only minimal, localized variations.

This comprehensive validation approach, combining quantitative classifier-based testing with dimensional visualization, provides strong confidence that the expanded dataset created through synthetic data generation maintains the statistical integrity needed for reliable counterfactual outcome modeling and bandit algorithm training, while addressing the original data scarcity issues without compromising clinical validity.

\subsection{Counterfactual Outcome Modeling Results}

The creation of precise counterfactual outcome models stands as a vital element for establishing effective treatment recommendation systems. Throughout this research, we leveraged a T-learner configuration to determine treatment effects by predicting potential outcomes across diverse treatment pathways. This section examines the performance metrics of various machine learning algorithms incorporated into our T-learner structure.

\subsubsection{Model Performance Comparison}

We examined four machine learning algorithms as base learners within the T-learner approach: Support Vector Machines (SVM), Random Forest, XGBoost, and Neural Networks. Each algorithm was developed to predict counterfactual outcomes for patients under varying treatment scenarios, following the T-learner framework that builds distinct models for each treatment condition. The performance metrics for each algorithm across all treatment groups are presented in Table \ref{tab:c_model_performance}. Model performance was evaluated using conventional prediction accuracy measures, including accuracy, precision, recall, and F1-score, which assessed the models' ability to predict observed outcomes on a reserved test dataset.

\begin{table}[htbp]
    \centering
        \caption{Performance Comparison of Machine Learning Algorithms in T-learner Framework}
        \vspace{6pt}
        \label{tab:c_model_performance}
        \begin{tabular}{lcccc}
            \hline
            \textbf{Algorithm} & \textbf{Accuracy (\%)} & \textbf{Precision (\%)} & \textbf{Recall (\%)} & \textbf{F1-Score (\%)} \\
            \hline
            XGBoost & 84.3 & 89.2 & 80.1 & 84.5 \\
            Random Forest & 82.6 & 86.7 & 77.0 & 81.6 \\
            Neural Network & 80.2 & 82.3 & 74.4 & 78.2 \\
            SVM & 79.3 & 80.5 & 72.8 & 76.4 \\
            \hline
        \end{tabular}
\end{table}

\subsubsection{Evaluation Methodology}

Our evaluation focused on direct prediction accuracy of potential outcomes, as this most directly impacts the subsequent bandit algorithm's ability to make optimal treatment decisions. We used 80\% of our combined original and synthetic dataset for training and the remaining 20\% for testing.

For each treatment arm, we trained separate predictors following the T-learner approach and evaluated how accurately they could predict the observed outcomes in cases where that treatment was actually administered. This allows us to assess the models' accuracy in the factual setting before relying on them for counterfactual estimates.

The accuracy percentage denotes the ratio of cases where the predicted outcome value aligned with the observed outcome within a clinically permissible error range. Complementary metrics such as precision, recall, and F1-score deliver a comprehensive analysis of model performance across multiple evaluation criteria.

\subsubsection{Model-Specific Insights}

XGBoost showed the best performance achieving an overall accuracy of 84.3\% across all treatment conditions. Its superior predictive capacity results from its adept processing of complex relationships between patient attributes and treatment effects, combined with its stability when handling the heterogeneous feature types present in our structured clinical data.

Random Forest ranked second in performance with an accuracy of 82.6\%. It showed remarkable aptitude in handling the categorical variables that represented clinical categorizations and discrete patient attributes.

The Neural Network model produced an accuracy of 80.2\%. Although neural networks have the capacity to learn complex non-linear relationships, the relatively limited dataset size may have restricted their performance effectiveness.

SVM models registered the poorest accuracy (79.3\%) compared to other algorithms tested, demonstrating limited ability to model the non-linear relationships between patient characteristics and treatment outcomes found within our clinical dataset. This performance differential between SVM and ensemble-based approaches reveals that treatment effect estimation requires capturing sophisticated interactions, which tree-based ensemble methods accomplish more effectively.

\subsubsection{Implications for Bandit-Based Treatment Recommendations}

The variations in model performance have significant ramifications for the implementation of the bandit algorithm that follows. XGBoost's enhanced predictive accuracy indicates it would yield the most trustworthy counterfactual estimates for directing optimal treatment allocation. Operating within the T-learner architecture to predict potential outcomes across different treatment alternatives, these performance benefits result in superior treatment effect estimation precision, thereby establishing the cornerstone of the contextual bandit reward structure.

Given these findings, we chose the XGBoost algorithm as the main model for generating counterfactual outcome predictions to underpin the bandit-driven treatment recommendation framework outlined in the following sections. The strong performance characteristics of this model guarantee that the bandit algorithms can depend on precise estimates of potential outcomes during treatment decision-making processes.

\subsection{Bandit Algorithms Training and Evaluation}

This section presents the results of training and evaluating various bandit algorithms for treatment recommendation, assessing their performance in optimizing patient outcomes based on the counterfactual models developed in the previous section.

\subsubsection{Bandit Algorithm Hyperparameter Optimization}

\begin{figure}[htbp]
    \centering
    \includegraphics[width=0.8\textwidth]{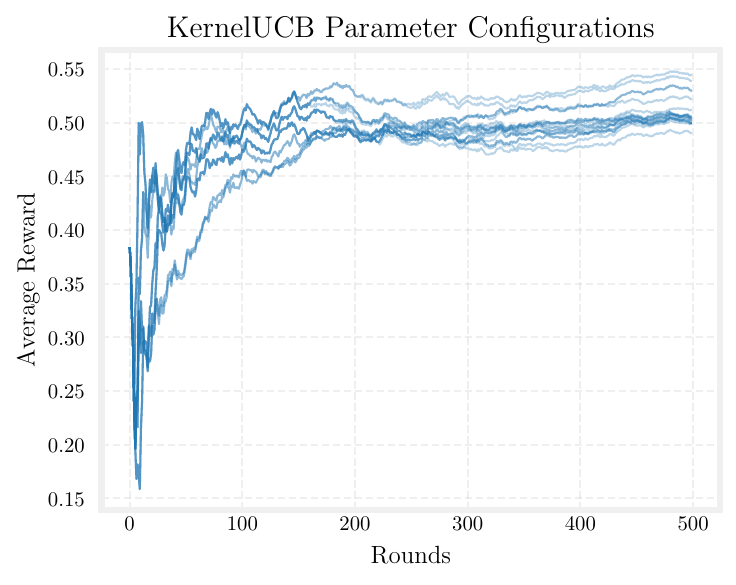}
    \caption{Hyperparameter tuning process for KernelBandit}
    \label{fig:tune-kb}
\end{figure}

We implemented thorough hyperparameter analysis for both KernelUCB and NeuralBandit contextual bandit methods. Figure \ref{fig:tune-kb} depicts KernelUCB's performance sensitivity across 20 configurations obtained via complete grid search over exploration parameter $\alpha \in \{0.1, 0.5\}$, kernel functions $\in \{$RBF, polynomial, linear$\}$, bandwidth $\gamma \in \{0.1, 0.5\}$ for RBF kernels, polynomial degree $\in \{2, 3\}$, regularization $\lambda_{reg} = 0.01$, and sample memory limits $\in \{100, 500\}$. The trajectories in Figure \ref{fig:tune-kb} show cumulative average rewards for each configuration throughout 500 rounds, exhibiting three distinguishable learning phases.

First, an initial exploration phase (rounds 0-20) where confidence intervals in the UCB optimism term cause nearly random arm sampling, producing volatile rewards occasionally dropping to $\approx$ 0.16. Hyperparameter choice has minimal influence during this initial evidence-gathering phase. Second, a rapid exploitation phase (rounds 20-120) where rewards climb steeply to 0.45-0.52 as initial posterior estimates form. Key patterns emerge during this phase: RBF kernels begin to dominate, confirming the underlying reward surface is smoothly non-linear; configurations with larger exploration coefficient ($\alpha = 0.5$) converge faster by encouraging broader early sampling; and configurations with larger replay memory (max\_samples = 500) gain advantage after approximately 80 rounds. Finally, an asymptotic plateau phase (rounds 120-500) where uncertainty shrinks and behavior becomes nearly greedy, with performance settling between 0.49 and 0.55. The optimal configuration—RBF kernel, $\alpha = 0.5$, $\lambda_{reg} = 0.01$, max\_samples = 500, $\gamma = 0.1$—maintains a consistent 0.03-0.05 performance advantage over the weakest settings (typically polynomial degree 3 with limited sample memory).

Our hyperparameter evaluation demonstrated that RBF kernels consistently surpassed both linear and polynomial variants, suggesting an underlying smooth, non-linear reward structure. A more restrictive bandwidth ($\gamma = 0.1$) more effectively captured local patterns while avoiding overfitting, whereas an elevated exploration coefficient altered the trade-off toward identifying optimal arms during initial phases, resulting in enhanced long-term performance despite early reward penalties.

\begin{figure}[htbp]
    \centering
    \includegraphics[width=0.8\textwidth]{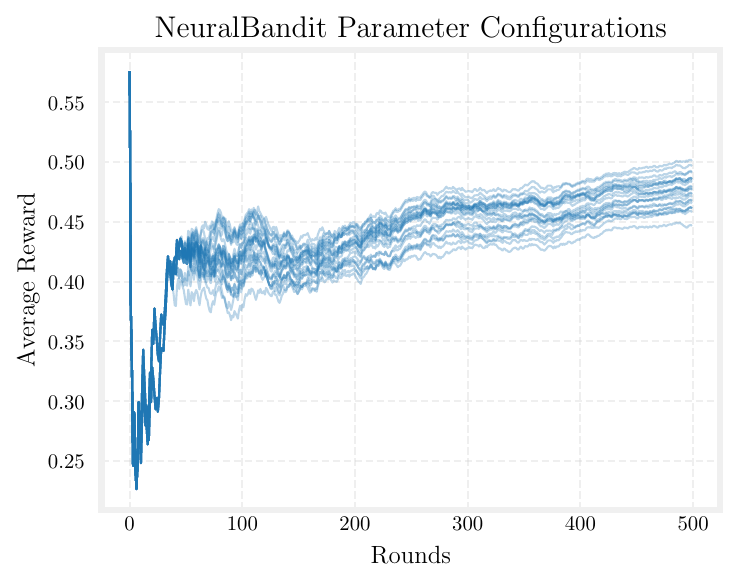}
    \caption{Hyperparameter tuning process for NeuralBandit}
    \label{fig:tune-nb}
\end{figure}

Figure \ref{fig:tune-nb} displays the hyperparameter tuning process for NeuralBandit, showing average rewards for 24 different configurations with parameters drawn from hidden\_size $\in \{32, 64, 128\}$, exploration parameter $\beta \in \{0.5, 1.0\}$, batch\_size $\in \{32, 64\}$, and learning-rate $\in \{10^{-3}, 10^{-2}\}$.

The NeuralBandit optimization curves demonstrate three learning phases: a warm-up period (rounds 0-15) characterized by near-random network weights and high exploration noise; a fast adaptation phase (rounds 15-120) where mini-batch gradient updates encode useful feature representations and push rewards toward 0.40-0.44; and a steady improvement phase (rounds 120-500) where agents converge to performance between 0.45 and 0.50.

Our examination demonstrated that expanded networks (hidden\_size = 128) and reduced batch sizes (32) exhibited accelerated learning through increased parameter updates per iteration. The superior NeuralBandit setup — hidden\_size = 128, $\beta = 0.5$, batch\_size = 32, learning-rate = 0.01 — attained a terminal reward of roughly 0.50, sustaining a 0.04-0.05 advantage over inferior configurations.

KernelUCB with RBF kernels outperformed NeuralBandit in our experiments, where the reward function was smooth and moderately low-dimensional. The RBF kernel's $\gamma = 0.1$ parameter proved well-suited to this structure, permitting closed-form Bayesian-like updates and faster convergence. NeuralBandit, however, had to learn the reward structure from scratch using gradient descent, which took longer and never matched KernelUCB's final performance during our evaluation period.

\subsubsection{Comparative Performance Evaluation}

After completing hyperparameter tuning, we performed an extensive comparative analysis of six treatment recommendation algorithms: Random policy (serving as our baseline), Epsilon-Greedy, UCB (Upper Confidence Bound), LinUCB, KernelUCB, and NeuralBandit. This evaluation utilized a simulation environment in which patient cases were produced through our validated CTGAN model, treatment recommendations were generated by each respective algorithm, and treatment outcomes were modeled using the XGBoost-based T-learner counterfactual framework.

\begin{figure}[htbp]
    \centering
    \includegraphics[width=0.8\textwidth]{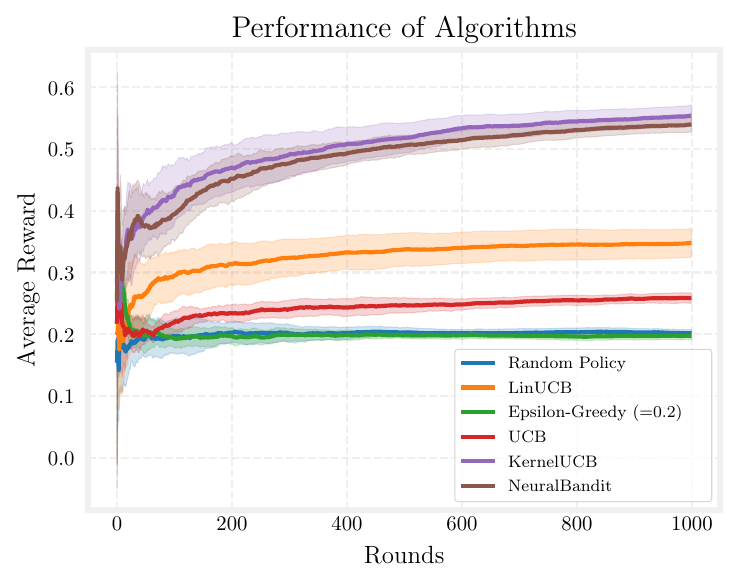}
    \caption{Comparative performance of bandit algorithms over 1,000 rounds}
    \label{fig:performance-1x}
\end{figure}

Figure \ref{fig:performance-1x} presents the learning curves of all six algorithms over 1,000 interaction rounds. Each curve represents the time-window-smoothed mean reward, with the surrounding ribbon denoting ±1 standard deviation computed from ten independent runs. All algorithms exhibit an initial phase of high variance and rapid growth during the first 30-50 rounds, where exploration dominates and reward estimates remain highly uncertain. As experience accumulates, the ribbons contract and the curves flatten, indicating convergence toward stable exploitation policies; by approximately round 400, additional gains become marginal for all algorithms.

KernelUCB shows the best overall results, reaching the maximum reward level ($\simeq 0.56$) with the steepest learning trajectory. This advantage comes from the kernel-enhanced Bayesian linear structure that can model intricate context-reward interactions while systematically diminishing uncertainty estimates. NeuralBandit performs reasonably well ($\simeq 0.53$) but exhibits wider fluctuation bands throughout the learning process, due to the probabilistic nature of gradient descent and stronger dependence on starting weight configurations.

LinUCB exhibits gradual improvement toward a moderate performance level ($\simeq 0.36$). Although its linear framework facilitates rapid learning when reward functions have linear characteristics, it struggles when non-linear patterns become prevalent in the reward structure. The context-free UCB method, which maintains optimistic exploration until uncertainty decreases, achieves roughly 0.26 with limited variance beyond 200 rounds, indicating stable performance once its confidence bounds converge.

By comparison, Epsilon-Greedy with fixed $\varepsilon = 0.20$ levels off at roughly 0.21 but keeps a wide ribbon throughout—since every fifth action is random, outcomes differ significantly across runs and performance suffers from constant exploration. The random policy stays predictably close to the overall arm mean ($\simeq 0.20$), with its ribbon shrinking as expected under the law of large numbers.

The empirical patterns suggest that the reward surface in our treatment recommendation task is highly non-linear but locally smooth—conditions that favor kernel methods over both deep and strictly linear alternatives. The KernelUCB algorithm with an RBF kernel exploits this structure exceptionally well: the RBF prior encodes the assumption that treatments with similar context vectors should yield similar rewards, allowing the posterior to quickly concentrate around promising regions of the context space.

\subsubsection{Long-Term Performance Analysis}

\begin{figure}[htbp]
    \centering
    \includegraphics[width=0.8\textwidth]{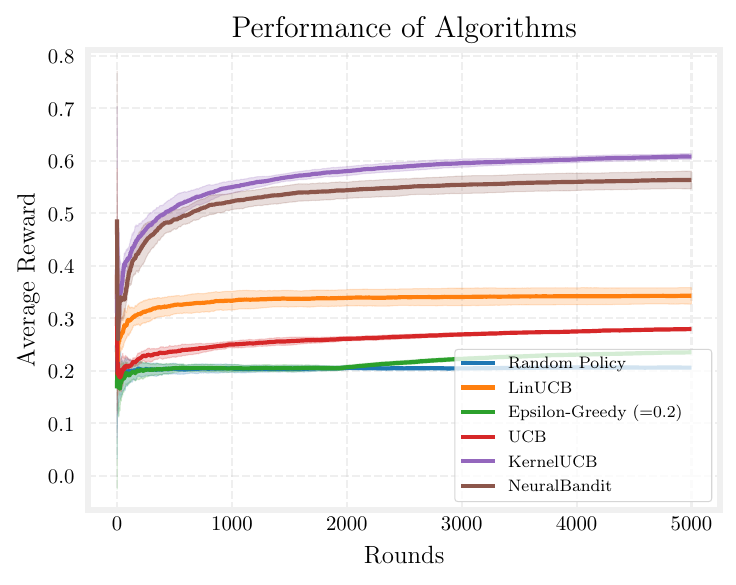}
    \caption{Comparative performance of bandit algorithms over 5,000 rounds}
    \label{fig:performance-5x}
\end{figure}

To assess the algorithms' performance trajectory over extended use, we conducted a longer-term evaluation over 5,000 recommendation rounds, illustrated in Figure \ref{fig:performance-5x}. This extended evaluation reveals several important patterns in both mean performance and variability while maintaining the same qualitative ordering of algorithms: KernelUCB > NeuralBandit > LinUCB > UCB > Epsilon-Greedy > Random.

KernelUCB maintains its upward trajectory past 1,000 rounds, stabilizing around 0.60-0.61 and capturing about four additional percentage points in mean reward. NeuralBandit experiences more substantial gains from the extended data, climbing to $\simeq 0.56$-$0.57$ and narrowing its performance difference with the kernel approach, though it doesn't catch up. This reveals that neural bandits require substantial sample sizes before their representational flexibility becomes worthwhile, whereas the RBF kernel's smoothness bias lets KernelUCB extract important structure from smaller datasets.

For LinUCB and the context-free baselines, the learning curves flatten much earlier, demonstrating diminishing returns for simpler models. LinUCB inches from $\simeq 0.36$ to $\simeq 0.35$-$0.36$, indicating that almost all linearly explainable signal was already exploited within the first thousand interactions. UCB makes a modest additional gain, stabilizing near 0.25-0.26, while the Epsilon-Greedy policy remains fixed at $\simeq 0.21$, its performance limited by the constant 20\% exploration rate regardless of horizon length. The random baseline, as expected, stays at $\simeq 0.20$, with its ribbon shrinking further under the law of large numbers.

The standard deviation ribbons are uniformly narrower than in the 1,000-round plot, reflecting lower run-to-run variance after prolonged learning. This contraction is most pronounced for KernelUCB and UCB, whose uncertainty-driven exploration schedules progressively eliminate epistemic variance. NeuralBandit retains a slightly wider band due to SGD stochasticity, while Epsilon-Greedy's ribbon levels off rather than vanishing because fresh randomness is injected every round.

These extended results confirm that KernelUCB with an RBF kernel provides the best trade-off between sample efficiency and asymptotic reward in our clinical treatment recommendation task. It not only reaches the highest performance but also converges quickly enough that additional interaction time yields only marginal improvements. Despite NeuralBandit's improved performance with extensive data, the kernel method maintains its advantage throughout 5,000 rounds. LinUCB's performance is intrinsically constrained by its linear modeling framework, and after a few hundred iterations, additional data cannot rectify the core model mismatch.

In conclusion, our assessment reveals that non-linear, context-sensitive approaches are superior for this treatment recommendation problem, where kernel-based methods perform optimally with limited data resources while deep learning models need extensive interaction to achieve their maximum capabilities. Basic linear or context-independent methods reach their performance plateau rapidly and fail to capture significant remaining potential.

\section{Conclusion}

This study introduced a comprehensive framework for optimizing treatment recommendations through a prior-informed method, utilizing bandit algorithms that were trained on historical data processed by large language models. The framework's practical utility and effectiveness were validated via a case study that concentrated on optimizing adjuvant chemotherapy protocols following surgery for stage-III colon cancer patients. Our main goal was to tackle the challenge of treatment personalization in clinical environments, especially when confronted with sparse observational datasets and the necessity to convert unstructured clinical documentation into actionable information. The approach combined LLM-driven data structuring, synthetic data generation through CTGAN, counterfactual outcome prediction using T-learners, and prior-informed contextual bandit methodologies.

Our results confirm the effectiveness of this multi-stage methodology. We determined that few-shot learning utilizing open-source LLMs, particularly DeepSeek-R1, can effectively structure narrative clinical documentation with considerable precision (93.2\% accuracy), surpassing performance of comparable models such as Llama 3.1. The subsequent application of CTGAN demonstrated success in expanding the structured dataset, producing synthetic patient records that accurately reflected the statistical characteristics of the original data, as confirmed through a two-sample test (AUC of 0.55) and t-SNE visualizations. In terms of counterfactual modeling, XGBoost proved to be the most reliable algorithm within the T-learner framework, attaining 84.3\% accuracy in forecasting treatment outcomes. This provided a solid foundation for simulating rewards during the bandit learning stage.

The central aim of our work involved assessing how different bandit algorithm approaches performed in practice. Through experimental comparison using simulated patient data, we determined that contextual bandits enhanced with prior domain knowledge dramatically outperformed traditional baseline methods. KernelUCB showed particularly impressive results, exploiting its kernel-based structure to recognize intricate non-linear relationships and benefiting from warm initialization with expert insights, which delivered excellent outcomes that rapidly secured an average reward of about 0.56 during the first 1,000 rounds and evolved to approximately 0.60-0.61 over the full 5,000-round assessment period.This performance exceeded both NeuralBandit ($\simeq$0.57 after 5,000 rounds) and LinUCB ($\simeq$0.36 after 5,000 rounds), demonstrating the critical value of incorporating non-linear modeling capabilities and leveraging prior knowledge effectively within this clinical domain. The outcomes reveal how online learning methodologies can progressively refine and enhance treatment decision-making processes while maintaining sample efficiency.

While these findings show promise, several limitations warrant consideration. First, the evaluations were performed entirely in-silico; prospective clinical validation is required prior to real-world implementation. Second, although the framework's applicability is demonstrated for stage-III colon cancer, its generalizability to other diseases with different data characteristics and reward structures needs further investigation. Third, although LLM-based few-shot learning for data extraction demonstrates effectiveness, it may not fully capture the breadth of clinical nuances that would be obtained through complete fine-tuning, and it carries the risk of incorporating biases present in the few-shot examples or the base LLM. There is also concern regarding the propagation of semantic errors from LLM extraction into the bandit priors. Finally, the single-objective reward function employed here represents a simplification of the complex, multi-criteria decision-making processes that characterize actual clinical practice.

Future research should emphasize prospective, adaptive medical recommendations to verify and improve these prior-informed bandit systems within actual clinical settings. These investigations should incorporate multi-dimensional reward structures that consider therapeutic effectiveness, adverse event profiles, financial implications, and patient well-being metrics across diverse patient cohorts. From a methodological perspective, LLM integration would benefit from research into interactive data extraction frameworks incorporating clinician-in-the-loop validation and approaches for measuring and transmitting uncertainty from LLM outputs to bandit priors. Regarding synthetic data generation, improving CTGANs through formal differential privacy protections and techniques that better maintain rare clinical subgroups represents a critical advancement. Exploring hybrid frameworks that combine offline reinforcement learning with online bandit adaptation may better leverage comprehensive historical data both before deployment and throughout active deployment phases. Furthermore, researching federated learning methodologies for CTGAN synthesis and bandit optimization across distributed healthcare organizations could enable privacy-preserving collaborative model improvement.

In conclusion, this research outlines a practical pathway for developing more responsive and individualized healthcare frameworks. Through the strategic integration of LLM-based knowledge extraction, synthetic data enhancement, rigorous counterfactual modeling, and systematic online learning via contextual bandits, we have established a workable methodology for creating evidence-based treatment recommendation systems.

\section*{CRediT authorship contribution statement}
\textbf{Saman Nessari:} Conceptualization, Methodology, Formal analysis, Software, Data curation, Writing – original draft, Writing – review \& editing.
\textbf{Ali Bozorgi-Amiri:} Conceptualization, Supervision, Resources, Validation, Writing – review \& editing.

\section*{Declaration of competing interest}
The authors declare that they have no competing interests.

\section*{Data availability}
The data that support the findings of this study are available on request from the corresponding author. The data are not publicly available due to privacy/ethical restrictions.

\section*{Acknowledgments}
The authors would like to express their sincere gratitude to all those who supported them throughout the course of this work. This project was completed without any external funding, relying solely on personal commitment and available resources.


\bibliography{literature.bib} 

\begin{thebibliography}{}

\bibitem [\protect \citeauthoryear {%
Akbasli%
, Birbilen%
\BCBL {}\ \BBA {} Teksam%
}{%
Akbasli%
\ \protect \BOthers {.}}{%
{\protect \APACyear {2025}}%
}]{%
akbasli2025leveraging}
\APACinsertmetastar {%
akbasli2025leveraging}%
\begin{APACrefauthors}%
Akbasli, I\BPBI T.%
, Birbilen, A\BPBI Z.%
\BCBL {}\ \BBA {} Teksam, O.%
\end{APACrefauthors}%
\unskip\
\newblock
\APACrefYearMonthDay{2025}{}{}.
\newblock
{\BBOQ}\APACrefatitle {Leveraging large language models to mimic domain expert
  labeling in unstructured text-based electronic healthcare records in
  non-english languages} {Leveraging large language models to mimic domain
  expert labeling in unstructured text-based electronic healthcare records in
  non-english languages}.{\BBCQ}
\newblock
\APACjournalVolNumPages{BMC Medical Informatics and Decision
  Making}{25}{1}{154}.
\PrintBackRefs{\CurrentBib}

\bibitem [\protect \citeauthoryear {%
Alamdari%
, Cao%
\BCBL {}\ \BBA {} Wilson%
}{%
Alamdari%
\ \protect \BOthers {.}}{%
{\protect \APACyear {2024}}%
}]{%
alamdari2024jump}
\APACinsertmetastar {%
alamdari2024jump}%
\begin{APACrefauthors}%
Alamdari, P\BPBI A.%
, Cao, Y.%
\BCBL {}\ \BBA {} Wilson, K\BPBI H.%
\end{APACrefauthors}%
\unskip\
\newblock
\APACrefYearMonthDay{2024}{}{}.
\newblock
{\BBOQ}\APACrefatitle {Jump starting bandits with llm-generated prior
  knowledge} {Jump starting bandits with llm-generated prior knowledge}.{\BBCQ}
\newblock
\APACjournalVolNumPages{arXiv preprint arXiv:2406.19317}{}{}{}.
\PrintBackRefs{\CurrentBib}

\bibitem [\protect \citeauthoryear {%
Alba%
, Xue%
, Abraham%
, Kannampallil%
\BCBL {}\ \BBA {} Lu%
}{%
Alba%
\ \protect \BOthers {.}}{%
{\protect \APACyear {2025}}%
}]{%
alba2025foundational}
\APACinsertmetastar {%
alba2025foundational}%
\begin{APACrefauthors}%
Alba, C.%
, Xue, B.%
, Abraham, J.%
, Kannampallil, T.%
\BCBL {}\ \BBA {} Lu, C.%
\end{APACrefauthors}%
\unskip\
\newblock
\APACrefYearMonthDay{2025}{}{}.
\newblock
{\BBOQ}\APACrefatitle {The foundational capabilities of large language models
  in predicting postoperative risks using clinical notes} {The foundational
  capabilities of large language models in predicting postoperative risks using
  clinical notes}.{\BBCQ}
\newblock
\APACjournalVolNumPages{npj Digital Medicine}{8}{1}{95}.
\PrintBackRefs{\CurrentBib}

\bibitem [\protect \citeauthoryear {%
Aslanyan%
\ \protect \BOthers {.}}{%
Aslanyan%
\ \protect \BOthers {.}}{%
{\protect \APACyear {2025}}%
}]{%
aslanyan2025bayesian}
\APACinsertmetastar {%
aslanyan2025bayesian}%
\begin{APACrefauthors}%
Aslanyan, V.%
, Pickering, T.%
, Nu{\~n}o, M.%
, Renfro, L\BPBI A.%
, Pa, J.%
\BCBL {}\ \BBA {} Mack, W\BPBI J.%
\end{APACrefauthors}%
\unskip\
\newblock
\APACrefYearMonthDay{2025}{}{}.
\newblock
{\BBOQ}\APACrefatitle {Bayesian Response Adaptive Randomization for Randomized
  Clinical Trials With Continuous Outcomes: The Role of Covariate Adjustment}
  {Bayesian response adaptive randomization for randomized clinical trials with
  continuous outcomes: The role of covariate adjustment}.{\BBCQ}
\newblock
\APACjournalVolNumPages{Pharmaceutical Statistics}{24}{2}{e2443}.
\PrintBackRefs{\CurrentBib}

\bibitem [\protect \citeauthoryear {%
Badreddine%
\ \BBA {} Spranger%
}{%
Badreddine%
\ \BBA {} Spranger%
}{%
{\protect \APACyear {2019}}%
}]{%
badreddine2019injecting}
\APACinsertmetastar {%
badreddine2019injecting}%
\begin{APACrefauthors}%
Badreddine, S.%
\BCBT {}\ \BBA {} Spranger, M.%
\end{APACrefauthors}%
\unskip\
\newblock
\APACrefYearMonthDay{2019}{}{}.
\newblock
{\BBOQ}\APACrefatitle {Injecting prior knowledge for transfer learning into
  reinforcement learning algorithms using logic tensor networks} {Injecting
  prior knowledge for transfer learning into reinforcement learning algorithms
  using logic tensor networks}.{\BBCQ}
\newblock
\APACjournalVolNumPages{arXiv preprint arXiv:1906.06576}{}{}{}.
\PrintBackRefs{\CurrentBib}

\bibitem [\protect \citeauthoryear {%
Berrevoets%
, Verboven%
\BCBL {}\ \BBA {} Verbeke%
}{%
Berrevoets%
\ \protect \BOthers {.}}{%
{\protect \APACyear {2022}}%
}]{%
berrevoets2022treatment}
\APACinsertmetastar {%
berrevoets2022treatment}%
\begin{APACrefauthors}%
Berrevoets, J.%
, Verboven, S.%
\BCBL {}\ \BBA {} Verbeke, W.%
\end{APACrefauthors}%
\unskip\
\newblock
\APACrefYearMonthDay{2022}{}{}.
\newblock
{\BBOQ}\APACrefatitle {Treatment effect optimisation in dynamic environments}
  {Treatment effect optimisation in dynamic environments}.{\BBCQ}
\newblock
\APACjournalVolNumPages{Journal of Causal Inference}{10}{1}{106--122}.
\PrintBackRefs{\CurrentBib}

\bibitem [\protect \citeauthoryear {%
Bhuyan%
\ \protect \BOthers {.}}{%
Bhuyan%
\ \protect \BOthers {.}}{%
{\protect \APACyear {2025}}%
}]{%
bhuyan2025generative}
\APACinsertmetastar {%
bhuyan2025generative}%
\begin{APACrefauthors}%
Bhuyan, S\BPBI S.%
, Sateesh, V.%
, Mukul, N.%
, Galvankar, A.%
, Mahmood, A.%
, Nauman, M.%
\BDBL {}Samuel, J.%
\end{APACrefauthors}%
\unskip\
\newblock
\APACrefYearMonthDay{2025}{}{}.
\newblock
{\BBOQ}\APACrefatitle {Generative Artificial Intelligence Use in Healthcare:
  Opportunities for Clinical Excellence and Administrative Efficiency}
  {Generative artificial intelligence use in healthcare: Opportunities for
  clinical excellence and administrative efficiency}.{\BBCQ}
\newblock
\APACjournalVolNumPages{Journal of Medical Systems}{49}{1}{10}.
\PrintBackRefs{\CurrentBib}

\bibitem [\protect \citeauthoryear {%
Chen%
\ \BBA {} Khademi%
}{%
Chen%
\ \BBA {} Khademi%
}{%
{\protect \APACyear {2024}}%
}]{%
chen2024adaptive}
\APACinsertmetastar {%
chen2024adaptive}%
\begin{APACrefauthors}%
Chen, N.%
\BCBT {}\ \BBA {} Khademi, A.%
\end{APACrefauthors}%
\unskip\
\newblock
\APACrefYearMonthDay{2024}{}{}.
\newblock
{\BBOQ}\APACrefatitle {Adaptive Seamless Dose-Finding Trials} {Adaptive
  seamless dose-finding trials}.{\BBCQ}
\newblock
\APACjournalVolNumPages{Manufacturing \& Service Operations
  Management}{26}{5}{1656--1673}.
\PrintBackRefs{\CurrentBib}

\bibitem [\protect \citeauthoryear {%
Chung%
}{%
Chung%
}{%
{\protect \APACyear {2025}}%
}]{%
chung2025artificial}
\APACinsertmetastar {%
chung2025artificial}%
\begin{APACrefauthors}%
Chung, D.%
\end{APACrefauthors}%
\unskip\
\newblock
\APACrefYearMonthDay{2025}{}{}.
\newblock
{\BBOQ}\APACrefatitle {Artificial intelligence in healthcare and medicine
  technology development review} {Artificial intelligence in healthcare and
  medicine technology development review}.{\BBCQ}
\newblock
\APACjournalVolNumPages{Engineering Applications of Artificial
  Intelligence}{143}{}{109801}.
\PrintBackRefs{\CurrentBib}

\bibitem [\protect \citeauthoryear {%
del Moral-Gonz{\'a}lez%
, G{\'o}mez-Adorno%
\BCBL {}\ \BBA {} Ramos-Flores%
}{%
del Moral-Gonz{\'a}lez%
\ \protect \BOthers {.}}{%
{\protect \APACyear {2025}}%
}]{%
del2025comparative}
\APACinsertmetastar {%
del2025comparative}%
\begin{APACrefauthors}%
del Moral-Gonz{\'a}lez, R.%
, G{\'o}mez-Adorno, H.%
\BCBL {}\ \BBA {} Ramos-Flores, O.%
\end{APACrefauthors}%
\unskip\
\newblock
\APACrefYearMonthDay{2025}{}{}.
\newblock
{\BBOQ}\APACrefatitle {Comparative analysis of generative LLMs for labeling
  entities in clinical notes} {Comparative analysis of generative llms for
  labeling entities in clinical notes}.{\BBCQ}
\newblock
\APACjournalVolNumPages{Genomics \& Informatics}{23}{}{3}.
\PrintBackRefs{\CurrentBib}

\bibitem [\protect \citeauthoryear {%
Esmaeilzadeh%
}{%
Esmaeilzadeh%
}{%
{\protect \APACyear {2024}}%
}]{%
esmaeilzadeh2024challenges}
\APACinsertmetastar {%
esmaeilzadeh2024challenges}%
\begin{APACrefauthors}%
Esmaeilzadeh, P.%
\end{APACrefauthors}%
\unskip\
\newblock
\APACrefYearMonthDay{2024}{}{}.
\newblock
{\BBOQ}\APACrefatitle {Challenges and strategies for wide-scale artificial
  intelligence (AI) deployment in healthcare practices: A perspective for
  healthcare organizations} {Challenges and strategies for wide-scale
  artificial intelligence (ai) deployment in healthcare practices: A
  perspective for healthcare organizations}.{\BBCQ}
\newblock
\APACjournalVolNumPages{Artificial Intelligence in Medicine}{151}{}{102861}.
\PrintBackRefs{\CurrentBib}

\bibitem [\protect \citeauthoryear {%
Goktas%
\ \BBA {} Grzybowski%
}{%
Goktas%
\ \BBA {} Grzybowski%
}{%
{\protect \APACyear {2025}}%
}]{%
goktas2025shaping}
\APACinsertmetastar {%
goktas2025shaping}%
\begin{APACrefauthors}%
Goktas, P.%
\BCBT {}\ \BBA {} Grzybowski, A.%
\end{APACrefauthors}%
\unskip\
\newblock
\APACrefYearMonthDay{2025}{}{}.
\newblock
{\BBOQ}\APACrefatitle {Shaping the future of healthcare: Ethical clinical
  challenges and pathways to trustworthy AI} {Shaping the future of healthcare:
  Ethical clinical challenges and pathways to trustworthy ai}.{\BBCQ}
\newblock
\APACjournalVolNumPages{Journal of Clinical Medicine}{14}{5}{1605}.
\PrintBackRefs{\CurrentBib}

\bibitem [\protect \citeauthoryear {%
Gu%
\ \protect \BOthers {.}}{%
Gu%
\ \protect \BOthers {.}}{%
{\protect \APACyear {2025}}%
}]{%
gu2025scalable}
\APACinsertmetastar {%
gu2025scalable}%
\begin{APACrefauthors}%
Gu, B.%
, Shao, V.%
, Liao, Z.%
, Carducci, V.%
, Brufau, S\BPBI R.%
, Yang, J.%
\BCBL {}\ \BBA {} Desai, R\BPBI J.%
\end{APACrefauthors}%
\unskip\
\newblock
\APACrefYearMonthDay{2025}{}{}.
\newblock
{\BBOQ}\APACrefatitle {Scalable information extraction from free text
  electronic health records using large language models} {Scalable information
  extraction from free text electronic health records using large language
  models}.{\BBCQ}
\newblock
\APACjournalVolNumPages{BMC Medical Research Methodology}{25}{1}{23}.
\PrintBackRefs{\CurrentBib}

\bibitem [\protect \citeauthoryear {%
Hsu%
\ \BBA {} Roberts%
}{%
Hsu%
\ \BBA {} Roberts%
}{%
{\protect \APACyear {2025}}%
}]{%
hsu2025leveraging}
\APACinsertmetastar {%
hsu2025leveraging}%
\begin{APACrefauthors}%
Hsu, E.%
\BCBT {}\ \BBA {} Roberts, K.%
\end{APACrefauthors}%
\unskip\
\newblock
\APACrefYearMonthDay{2025}{}{}.
\newblock
{\BBOQ}\APACrefatitle {Leveraging large language models for knowledge-free weak
  supervision in clinical natural language processing} {Leveraging large
  language models for knowledge-free weak supervision in clinical natural
  language processing}.{\BBCQ}
\newblock
\APACjournalVolNumPages{Scientific Reports}{15}{1}{8241}.
\PrintBackRefs{\CurrentBib}

\bibitem [\protect \citeauthoryear {%
Huang%
, Niazmand%
\BCBL {}\ \BBA {} Vidal%
}{%
Huang%
\ \protect \BOthers {.}}{%
{\protect \APACyear {2024}}%
}]{%
huang2024hybrid}
\APACinsertmetastar {%
huang2024hybrid}%
\begin{APACrefauthors}%
Huang, H.%
, Niazmand, E.%
\BCBL {}\ \BBA {} Vidal, M\BHBI E.%
\end{APACrefauthors}%
\unskip\
\newblock
\APACrefYearMonthDay{2024}{}{}.
\newblock
{\BBOQ}\APACrefatitle {Hybrid AI Approach for Counterfactual Prediction over
  Knowledge Graphs for Personal Healthcare} {Hybrid ai approach for
  counterfactual prediction over knowledge graphs for personal
  healthcare}.{\BBCQ}
\newblock
\BIn{} \APACrefbtitle {Artificial Intelligence and Data Science for Healthcare:
  Bridging Data-Centric AI and People-Centric Healthcare.} {Artificial
  intelligence and data science for healthcare: Bridging data-centric ai and
  people-centric healthcare.}
\PrintBackRefs{\CurrentBib}

\bibitem [\protect \citeauthoryear {%
Jayaraman%
, Desman%
, Sabounchi%
, Nadkarni%
\BCBL {}\ \BBA {} Sakhuja%
}{%
Jayaraman%
\ \protect \BOthers {.}}{%
{\protect \APACyear {2024}}%
}]{%
jayaraman2024primer}
\APACinsertmetastar {%
jayaraman2024primer}%
\begin{APACrefauthors}%
Jayaraman, P.%
, Desman, J.%
, Sabounchi, M.%
, Nadkarni, G\BPBI N.%
\BCBL {}\ \BBA {} Sakhuja, A.%
\end{APACrefauthors}%
\unskip\
\newblock
\APACrefYearMonthDay{2024}{}{}.
\newblock
{\BBOQ}\APACrefatitle {A Primer on Reinforcement Learning in Medicine for
  Clinicians} {A primer on reinforcement learning in medicine for
  clinicians}.{\BBCQ}
\newblock
\APACjournalVolNumPages{NPJ Digital Medicine}{7}{1}{337}.
\PrintBackRefs{\CurrentBib}

\bibitem [\protect \citeauthoryear {%
Kumar%
, Chauhan%
\BCBL {}\ \BBA {} Awasthi%
}{%
Kumar%
\ \protect \BOthers {.}}{%
{\protect \APACyear {2023}}%
}]{%
kumar2023artificial}
\APACinsertmetastar {%
kumar2023artificial}%
\begin{APACrefauthors}%
Kumar, P.%
, Chauhan, S.%
\BCBL {}\ \BBA {} Awasthi, L\BPBI K.%
\end{APACrefauthors}%
\unskip\
\newblock
\APACrefYearMonthDay{2023}{}{}.
\newblock
{\BBOQ}\APACrefatitle {Artificial intelligence in healthcare: review, ethics,
  trust challenges \& future research directions} {Artificial intelligence in
  healthcare: review, ethics, trust challenges \& future research
  directions}.{\BBCQ}
\newblock
\APACjournalVolNumPages{Engineering Applications of Artificial
  Intelligence}{120}{}{105894}.
\PrintBackRefs{\CurrentBib}

\bibitem [\protect \citeauthoryear {%
Kyrimi%
\ \protect \BOthers {.}}{%
Kyrimi%
\ \protect \BOthers {.}}{%
{\protect \APACyear {2025}}%
}]{%
kyrimi2025counterfactual}
\APACinsertmetastar {%
kyrimi2025counterfactual}%
\begin{APACrefauthors}%
Kyrimi, E.%
, Mossadegh, S.%
, Wohlgemut, J\BPBI M.%
, Stoner, R\BPBI S.%
, Tai, N\BPBI R.%
\BCBL {}\ \BBA {} Marsh, W.%
\end{APACrefauthors}%
\unskip\
\newblock
\APACrefYearMonthDay{2025}{}{}.
\newblock
{\BBOQ}\APACrefatitle {Counterfactual reasoning using causal Bayesian networks
  as a healthcare governance tool} {Counterfactual reasoning using causal
  bayesian networks as a healthcare governance tool}.{\BBCQ}
\newblock
\APACjournalVolNumPages{International journal of medical
  informatics}{193}{}{105681}.
\PrintBackRefs{\CurrentBib}

\bibitem [\protect \citeauthoryear {%
Liberali%
\ \protect \BOthers {.}}{%
Liberali%
\ \protect \BOthers {.}}{%
{\protect \APACyear {2025}}%
}]{%
liberali2025real}
\APACinsertmetastar {%
liberali2025real}%
\begin{APACrefauthors}%
Liberali, G.%
, Boersma, E.%
, Lingsma, H.%
, Brugts, J.%
, Dippel, D.%
, Tijssen, J.%
\BCBL {}\ \BBA {} Hauser, J.%
\end{APACrefauthors}%
\unskip\
\newblock
\APACrefYearMonthDay{2025}{}{}.
\newblock
{\BBOQ}\APACrefatitle {Real-time adaptive randomization of clinical trials}
  {Real-time adaptive randomization of clinical trials}.{\BBCQ}
\newblock
\APACjournalVolNumPages{Journal of Clinical Epidemiology}{178}{}{111612}.
\PrintBackRefs{\CurrentBib}

\bibitem [\protect \citeauthoryear {%
Lopez%
\ \protect \BOthers {.}}{%
Lopez%
\ \protect \BOthers {.}}{%
{\protect \APACyear {2025}}%
}]{%
lopez2025clinical}
\APACinsertmetastar {%
lopez2025clinical}%
\begin{APACrefauthors}%
Lopez, I.%
, Swaminathan, A.%
, Vedula, K.%
, Narayanan, S.%
, Nateghi~Haredasht, F.%
, Ma, S\BPBI P.%
\BDBL {}Chen, J\BPBI H.%
\end{APACrefauthors}%
\unskip\
\newblock
\APACrefYearMonthDay{2025}{}{}.
\newblock
{\BBOQ}\APACrefatitle {Clinical entity augmented retrieval for clinical
  information extraction} {Clinical entity augmented retrieval for clinical
  information extraction}{\BBCQ}\ [Article].
\newblock
\APACjournalVolNumPages{npj Digital Medicine}{8}{1}{}.
\newblock
\APACrefnote{Cited by: 3; All Open Access, Gold Open Access}
\PrintBackRefs{\CurrentBib}

\bibitem [\protect \citeauthoryear {%
Norwood%
, Davidian%
\BCBL {}\ \BBA {} Laber%
}{%
Norwood%
\ \protect \BOthers {.}}{%
{\protect \APACyear {2024}}%
}]{%
norwood2024adaptive}
\APACinsertmetastar {%
norwood2024adaptive}%
\begin{APACrefauthors}%
Norwood, P.%
, Davidian, M.%
\BCBL {}\ \BBA {} Laber, E.%
\end{APACrefauthors}%
\unskip\
\newblock
\APACrefYearMonthDay{2024}{}{}.
\newblock
{\BBOQ}\APACrefatitle {Adaptive randomization methods for sequential multiple
  assignment randomized trials (smarts) via thompson sampling} {Adaptive
  randomization methods for sequential multiple assignment randomized trials
  (smarts) via thompson sampling}.{\BBCQ}
\newblock
\APACjournalVolNumPages{Biometrics}{80}{4}{ujae152}.
\PrintBackRefs{\CurrentBib}

\bibitem [\protect \citeauthoryear {%
Oh%
, Park%
, Lee%
, Kang%
\BCBL {}\ \BBA {} Mo%
}{%
Oh%
\ \protect \BOthers {.}}{%
{\protect \APACyear {2022}}%
}]{%
oh2022reinforcement}
\APACinsertmetastar {%
oh2022reinforcement}%
\begin{APACrefauthors}%
Oh, S\BPBI H.%
, Park, J.%
, Lee, S\BPBI J.%
, Kang, S.%
\BCBL {}\ \BBA {} Mo, J.%
\end{APACrefauthors}%
\unskip\
\newblock
\APACrefYearMonthDay{2022}{}{}.
\newblock
{\BBOQ}\APACrefatitle {Reinforcement learning-based expanded personalized
  diabetes treatment recommendation using South Korean electronic health
  records} {Reinforcement learning-based expanded personalized diabetes
  treatment recommendation using south korean electronic health
  records}.{\BBCQ}
\newblock
\APACjournalVolNumPages{Expert Systems with Applications}{206}{}{117932}.
\PrintBackRefs{\CurrentBib}

\bibitem [\protect \citeauthoryear {%
Parvin%
, Chessa%
, Kaptein%
\BCBL {}\ \BBA {} Patern{\`o}%
}{%
Parvin%
\ \protect \BOthers {.}}{%
{\protect \APACyear {2019}}%
}]{%
parvin2019personalized}
\APACinsertmetastar {%
parvin2019personalized}%
\begin{APACrefauthors}%
Parvin, P.%
, Chessa, S.%
, Kaptein, M.%
\BCBL {}\ \BBA {} Patern{\`o}, F.%
\end{APACrefauthors}%
\unskip\
\newblock
\APACrefYearMonthDay{2019}{}{}.
\newblock
{\BBOQ}\APACrefatitle {Personalized real-time anomaly detection and health
  feedback for older adults} {Personalized real-time anomaly detection and
  health feedback for older adults}.{\BBCQ}
\newblock
\APACjournalVolNumPages{Journal of ambient intelligence and smart
  environments}{11}{5}{453--469}.
\PrintBackRefs{\CurrentBib}

\bibitem [\protect \citeauthoryear {%
Pasquadibisceglie%
, Appice%
, Malerba%
\BCBL {}\ \BBA {} Fiameni%
}{%
Pasquadibisceglie%
\ \protect \BOthers {.}}{%
{\protect \APACyear {2025}}%
}]{%
pasquadibisceglie2025leveraging}
\APACinsertmetastar {%
pasquadibisceglie2025leveraging}%
\begin{APACrefauthors}%
Pasquadibisceglie, V.%
, Appice, A.%
, Malerba, D.%
\BCBL {}\ \BBA {} Fiameni, G.%
\end{APACrefauthors}%
\unskip\
\newblock
\APACrefYearMonthDay{2025}{}{}.
\newblock
{\BBOQ}\APACrefatitle {Leveraging a Large Language Model (LLM) to Predict
  Hospital Admissions of Emergency Department Patients} {Leveraging a large
  language model (llm) to predict hospital admissions of emergency department
  patients}.{\BBCQ}
\newblock
\APACjournalVolNumPages{Expert Systems with Applications}{}{}{128224}.
\PrintBackRefs{\CurrentBib}

\bibitem [\protect \citeauthoryear {%
Prosperi%
\ \protect \BOthers {.}}{%
Prosperi%
\ \protect \BOthers {.}}{%
{\protect \APACyear {2020}}%
}]{%
prosperi2020causal}
\APACinsertmetastar {%
prosperi2020causal}%
\begin{APACrefauthors}%
Prosperi, M.%
, Guo, Y.%
, Sperrin, M.%
, Koopman, J.%
, Min, J.%
, He, X.%
\BDBL {}Bian, J.%
\end{APACrefauthors}%
\unskip\
\newblock
\APACrefYearMonthDay{2020}{}{}.
\newblock
\APACrefbtitle {Causal inference and counterfactual prediction in machine
  learning for actionable healthcare. Nat Mach Intell 2: 369--375.} {Causal
  inference and counterfactual prediction in machine learning for actionable
  healthcare. nat mach intell 2: 369--375.}
\PrintBackRefs{\CurrentBib}

\bibitem [\protect \citeauthoryear {%
Qian%
, Wang%
\BCBL {}\ \BBA {} Zhao%
}{%
Qian%
\ \protect \BOthers {.}}{%
{\protect \APACyear {2025}}%
}]{%
qian2025personalized}
\APACinsertmetastar {%
qian2025personalized}%
\begin{APACrefauthors}%
Qian, S.%
, Wang, J.%
\BCBL {}\ \BBA {} Zhao, S.%
\end{APACrefauthors}%
\unskip\
\newblock
\APACrefYearMonthDay{2025}{}{}.
\newblock
{\BBOQ}\APACrefatitle {A personalized active learning strategy with enhanced
  user satisfaction for recommender systems} {A personalized active learning
  strategy with enhanced user satisfaction for recommender systems}.{\BBCQ}
\newblock
\APACjournalVolNumPages{Expert Systems with Applications}{}{}{128765}.
\PrintBackRefs{\CurrentBib}

\bibitem [\protect \citeauthoryear {%
Varatharajah%
\ \BBA {} Berry%
}{%
Varatharajah%
\ \BBA {} Berry%
}{%
{\protect \APACyear {2022}}%
}]{%
varatharajah2022contextual}
\APACinsertmetastar {%
varatharajah2022contextual}%
\begin{APACrefauthors}%
Varatharajah, Y.%
\BCBT {}\ \BBA {} Berry, B.%
\end{APACrefauthors}%
\unskip\
\newblock
\APACrefYearMonthDay{2022}{}{}.
\newblock
{\BBOQ}\APACrefatitle {A contextual-bandit-based approach for informed
  decision-making in clinical trials} {A contextual-bandit-based approach for
  informed decision-making in clinical trials}.{\BBCQ}
\newblock
\APACjournalVolNumPages{Life}{12}{8}{1277}.
\PrintBackRefs{\CurrentBib}

\bibitem [\protect \citeauthoryear {%
Villar%
, Bowden%
\BCBL {}\ \BBA {} Wason%
}{%
Villar%
\ \protect \BOthers {.}}{%
{\protect \APACyear {2015}}%
}]{%
villar2015multi}
\APACinsertmetastar {%
villar2015multi}%
\begin{APACrefauthors}%
Villar, S\BPBI S.%
, Bowden, J.%
\BCBL {}\ \BBA {} Wason, J.%
\end{APACrefauthors}%
\unskip\
\newblock
\APACrefYearMonthDay{2015}{}{}.
\newblock
{\BBOQ}\APACrefatitle {Multi-armed bandit models for the optimal design of
  clinical trials: benefits and challenges} {Multi-armed bandit models for the
  optimal design of clinical trials: benefits and challenges}.{\BBCQ}
\newblock
\APACjournalVolNumPages{Statistical science: a review journal of the Institute
  of Mathematical Statistics}{30}{2}{199}.
\PrintBackRefs{\CurrentBib}

\bibitem [\protect \citeauthoryear {%
Wei%
, Ma%
\BCBL {}\ \BBA {} Wang%
}{%
Wei%
\ \protect \BOthers {.}}{%
{\protect \APACyear {2025}}%
}]{%
wei2025adaptive}
\APACinsertmetastar {%
wei2025adaptive}%
\begin{APACrefauthors}%
Wei, W.%
, Ma, X.%
\BCBL {}\ \BBA {} Wang, J.%
\end{APACrefauthors}%
\unskip\
\newblock
\APACrefYearMonthDay{2025}{}{}.
\newblock
{\BBOQ}\APACrefatitle {Adaptive experiments toward learning treatment effect
  heterogeneity} {Adaptive experiments toward learning treatment effect
  heterogeneity}.{\BBCQ}
\newblock
\APACjournalVolNumPages{Journal of the Royal Statistical Society Series B:
  Statistical Methodology}{}{}{qkaf006}.
\PrintBackRefs{\CurrentBib}

\bibitem [\protect \citeauthoryear {%
Wu%
, Shi%
, Choudhary%
\BCBL {}\ \BBA {} Wang%
}{%
Wu%
\ \protect \BOthers {.}}{%
{\protect \APACyear {2024}}%
}]{%
wu2024clinical}
\APACinsertmetastar {%
wu2024clinical}%
\begin{APACrefauthors}%
Wu, H.%
, Shi, W.%
, Choudhary, A.%
\BCBL {}\ \BBA {} Wang, M\BPBI D.%
\end{APACrefauthors}%
\unskip\
\newblock
\APACrefYearMonthDay{2024}{}{}.
\newblock
{\BBOQ}\APACrefatitle {Clinical decision making under uncertainty: a
  bootstrapped counterfactual inference approach} {Clinical decision making
  under uncertainty: a bootstrapped counterfactual inference approach}.{\BBCQ}
\newblock
\APACjournalVolNumPages{BMC Medical Informatics and Decision
  Making}{24}{1}{275}.
\PrintBackRefs{\CurrentBib}

\bibitem [\protect \citeauthoryear {%
Xian%
\ \protect \BOthers {.}}{%
Xian%
\ \protect \BOthers {.}}{%
{\protect \APACyear {2024}}%
}]{%
xian2024language}
\APACinsertmetastar {%
xian2024language}%
\begin{APACrefauthors}%
Xian, S.%
, Grabowska, M\BPBI E.%
, Kullo, I\BPBI J.%
, Luo, Y.%
, Smoller, J\BPBI W.%
, Wei, W\BHBI Q.%
\BDBL {}Crosslin, D.%
\end{APACrefauthors}%
\unskip\
\newblock
\APACrefYearMonthDay{2024}{}{}.
\newblock
{\BBOQ}\APACrefatitle {Language-model-based patient embedding using electronic
  health records facilitates phenotyping, disease forecasting, and progression
  analysis} {Language-model-based patient embedding using electronic health
  records facilitates phenotyping, disease forecasting, and progression
  analysis}.{\BBCQ}
\newblock
\APACjournalVolNumPages{Research Square}{}{}{rs--3}.
\PrintBackRefs{\CurrentBib}

\bibitem [\protect \citeauthoryear {%
Xie%
\ \protect \BOthers {.}}{%
Xie%
\ \protect \BOthers {.}}{%
{\protect \APACyear {2025}}%
}]{%
xie2025medical}
\APACinsertmetastar {%
xie2025medical}%
\begin{APACrefauthors}%
Xie, Q.%
, Chen, Q.%
, Chen, A.%
, Peng, C.%
, Hu, Y.%
, Lin, F.%
\BDBL {}Bian, J.%
\end{APACrefauthors}%
\unskip\
\newblock
\APACrefYearMonthDay{2025}{}{}.
\newblock
{\BBOQ}\APACrefatitle {Medical foundation large language models for
  comprehensive text analysis and beyond} {Medical foundation large language
  models for comprehensive text analysis and beyond}{\BBCQ}\ [Article].
\newblock
\APACjournalVolNumPages{npj Digital Medicine}{8}{1}{}.
\newblock
\APACrefnote{Cited by: 1}
\PrintBackRefs{\CurrentBib}

\bibitem [\protect \citeauthoryear {%
Xu%
, Chen%
, Hu%
\BCBL {}\ \BBA {} Li%
}{%
Xu%
\ \protect \BOthers {.}}{%
{\protect \APACyear {2025}}%
}]{%
xu2025staf}
\APACinsertmetastar {%
xu2025staf}%
\begin{APACrefauthors}%
Xu, T.%
, Chen, L.%
, Hu, Z.%
\BCBL {}\ \BBA {} Li, B.%
\end{APACrefauthors}%
\unskip\
\newblock
\APACrefYearMonthDay{2025}{}{}.
\newblock
{\BBOQ}\APACrefatitle {STAF-LLM: A scalable and task-adaptive fine-tuning
  framework for large language models in medical domain} {Staf-llm: A scalable
  and task-adaptive fine-tuning framework for large language models in medical
  domain}.{\BBCQ}
\newblock
\APACjournalVolNumPages{Expert Systems with Applications}{281}{}{127582}.
\PrintBackRefs{\CurrentBib}

\bibitem [\protect \citeauthoryear {%
Yankeelov%
\ \protect \BOthers {.}}{%
Yankeelov%
\ \protect \BOthers {.}}{%
{\protect \APACyear {2024}}%
}]{%
yankeelov2024designing}
\APACinsertmetastar {%
yankeelov2024designing}%
\begin{APACrefauthors}%
Yankeelov, T\BPBI E.%
, Hormuth, D\BPBI A.%
, Lima, E\BPBI A.%
, Lorenzo, G.%
, Wu, C.%
, Okereke, L\BPBI C.%
\BDBL {}Chung, C.%
\end{APACrefauthors}%
\unskip\
\newblock
\APACrefYearMonthDay{2024}{}{}.
\newblock
{\BBOQ}\APACrefatitle {Designing clinical trials for patients who are not
  average} {Designing clinical trials for patients who are not average}.{\BBCQ}
\newblock
\APACjournalVolNumPages{Iscience}{27}{1}{}.
\PrintBackRefs{\CurrentBib}

\bibitem [\protect \citeauthoryear {%
Zhalechian%
, Keyvanshokooh%
, Shi%
\BCBL {}\ \BBA {} Van~Oyen%
}{%
Zhalechian%
\ \protect \BOthers {.}}{%
{\protect \APACyear {2022}}%
}]{%
zhalechian2022online}
\APACinsertmetastar {%
zhalechian2022online}%
\begin{APACrefauthors}%
Zhalechian, M.%
, Keyvanshokooh, E.%
, Shi, C.%
\BCBL {}\ \BBA {} Van~Oyen, M\BPBI P.%
\end{APACrefauthors}%
\unskip\
\newblock
\APACrefYearMonthDay{2022}{}{}.
\newblock
{\BBOQ}\APACrefatitle {Online resource allocation with personalized learning}
  {Online resource allocation with personalized learning}.{\BBCQ}
\newblock
\APACjournalVolNumPages{Operations Research}{70}{4}{2138--2161}.
\PrintBackRefs{\CurrentBib}

\bibitem [\protect \citeauthoryear {%
P.~Zhang%
, Shi%
\BCBL {}\ \BBA {} Kamel~Boulos%
}{%
P.~Zhang%
\ \protect \BOthers {.}}{%
{\protect \APACyear {2024}}%
}]{%
zhang2024generative}
\APACinsertmetastar {%
zhang2024generative}%
\begin{APACrefauthors}%
Zhang, P.%
, Shi, J.%
\BCBL {}\ \BBA {} Kamel~Boulos, M\BPBI N.%
\end{APACrefauthors}%
\unskip\
\newblock
\APACrefYearMonthDay{2024}{}{}.
\newblock
{\BBOQ}\APACrefatitle {Generative ai in medicine and healthcare: Moving beyond
  the ‘peak of inflated expectations’} {Generative ai in medicine and
  healthcare: Moving beyond the ‘peak of inflated expectations’}.{\BBCQ}
\newblock
\APACjournalVolNumPages{Future Internet}{16}{12}{462}.
\PrintBackRefs{\CurrentBib}

\bibitem [\protect \citeauthoryear {%
Z.~Zhang%
\ \BBA {} Ni%
}{%
Z.~Zhang%
\ \BBA {} Ni%
}{%
{\protect \APACyear {2025}}%
}]{%
zhang2025critical}
\APACinsertmetastar {%
zhang2025critical}%
\begin{APACrefauthors}%
Zhang, Z.%
\BCBT {}\ \BBA {} Ni, H.%
\end{APACrefauthors}%
\unskip\
\newblock
\APACrefYearMonthDay{2025}{}{}.
\newblock
{\BBOQ}\APACrefatitle {Critical care studies using large language models based
  on electronic healthcare records: A technical note} {Critical care studies
  using large language models based on electronic healthcare records: A
  technical note}.{\BBCQ}
\newblock
\APACjournalVolNumPages{Journal of Intensive Medicine}{5}{02}{137--150}.
\PrintBackRefs{\CurrentBib}

\end{thebibliography}
\bibliographystyle{apacite} 

\end{document}